\documentclass[preprint]{elsarticle}

\usepackage{graphicx} 
\usepackage[table]{xcolor} 

\usepackage{amssymb}
\usepackage[utf8]{inputenc}
\usepackage{newunicodechar}
\newunicodechar{≈}{\approx}
\usepackage{amsmath}
\usepackage{amsthm}

\usepackage[all]{nowidow}

\usepackage{booktabs}

\usepackage{hyperref}

\usepackage{listings}

\usepackage{makecell}
\usepackage{multirow}

\usepackage{tabularx} 
\newcolumntype{Y}{>{\centering\arraybackslash}X}

\journal{arXiv}

\begin{document}

\begin{frontmatter}

\title{Using Large Language Models for Legal Decision-Making in Austrian Value-Added Tax Law: A Comparative Study}

\author[label1,label2]{Marina Luketina\corref{cor1}}
\ead{marina.luketina@fh-steyr.at}
\cortext[cor1]{Corresponding author}
\affiliation[label1]{
  organization={University of Applied Sciences Upper Austria -- Campus Steyr},
  addressline={Wehrgrabengasse~1--3},
  city={Steyr},
  postcode={4400},
  country={Austria}
}
\affiliation[label2]{
  organization={Johannes Kepler University Linz},
  addressline={Altenberger Str.~69},
  city={Linz},
  postcode={4040},
  country={Austria}
}

\author[label1,label2]{Andrea Benkel}
\ead{benkel@dke.uni-linz.ac.at}

\author[label2]{Christoph G. Schuetz}
\ead{schuetz@dke.uni-linz.ac.at}

\begin{abstract}
This paper provides an experimental evaluation of the capability of large language models (LLMs) to assist in legal decision-making within the framework of Austrian and European Union value-added tax (VAT) law. In tax consulting practice, clients often describe cases in natural language, making LLMs a prime candidate for supporting automated decision-making and reducing the workload of tax professionals. Given the requirement for legally grounded and well-justified analyses, the propensity of LLMs to hallucinate presents a considerable challenge. The experiments focus on two common methods for enhancing LLM performance: fine-tuning and retrieval-augmented generation (RAG).  In this study, these methods are applied on both textbook cases and real-world cases from a tax consulting firm to systematically determine the best configurations of LLM-based systems and assess the legal-reasoning capabilities of LLMs. The findings highlight the potential of using LLMs to support tax consultants by automating routine tasks and providing initial analyses, although current prototypes are not ready for full automation due to the sensitivity of the legal domain. The findings indicate that LLMs, when properly configured, can effectively support tax professionals in VAT tasks and provide legally grounded justifications for decisions. However, limitations remain regarding the handling of implicit client knowledge and context-specific documentation, underscoring the need for future integration of structured background information.
\end{abstract}

\begin{keyword}

applied artificial intelligence \sep fine-tuning \sep retrieval-augmented generation \sep tax consulting \sep legal reasoning

\end{keyword}

\end{frontmatter}

\section{Introduction}

Generative artificial intelligence (GenAI) is being increasingly developed to support knowledge workers in addressing complex, domain-specific challenges by simulating human-like reasoning.
Consequently, the role of artificial intelligence (AI) is shifting from automating routine tasks to acting as an advanced assistant in specialized fields such as medicine, auditing, law, and accounting. Central to this development are large language models (LLMs), which have become a major focus of research aimed at creating AI systems tailored to professional domains. 
As noted by Sako~\cite{sako_how_2024}, this trend is also evident in the legal sector, where research is growing around the use of AI to interpret and analyze complex legal texts~\cite{homoki_large_2024}.

The legal domain remains a particularly sensitive field for AI deployment due to its high demands for precision, compliance, and ethical integrity. 
This is especially true in tax law, where errors in legal interpretation can result in significant financial and legal consequences~\cite{alarie_rise_2023}. 
Within this field, value-added tax (VAT) law presents a compelling application for LLMs: VAT affects nearly all economic transactions and is harmonized across EU Member States~\cite{european_comission_value-added_2024}. 
This harmonization implies that an AI system developed for one Member State's VAT law could, in principle, be adapted for use across the EU.
A central challenge in VAT law lies in determining the place of supply of goods and services, as this defines the jurisdiction entitled to levy VAT.
Businesses must navigate these determinations daily to ensure compliance, as VAT is applied at every stage of the value chain and its exemptions or obligations must be legally substantiated. 
Given its complexity and practical relevance, VAT law offers a distinct test case for assessing the capacity of GenAI and LLMs to support legal reasoning and assist businesses with routine compliance challenges.

LLMs are general-purpose tools, not at all specialized on analyzing VAT cases, often trained on predominantly English-language documents.
To explore the potential of LLMs to support legal reasoning and assist businesses in navigating complex VAT compliance tasks, this study examines the effectiveness of two prominent adaptation methods for LLMs: fine-tuning and retrieval-augmented generation (RAG).
These techniques are first applied to enhance LLM performance on a special VAT-related question---namely, the determination of the place of supply or service provision under Austrian and EU VAT law. 
The evaluation focuses both on the accuracy of the generated responses and the quality of their legal reasoning, i.e., the explanations provided for the responses. 
With the models adapted to this narrowly defined legal task, their capabilities are then tested using real-world VAT cases. 
This two-step approach allows for a targeted improvement of the models' legal reasoning, followed by a realistic assessment of their accuracy and reliability in complex, practice-oriented scenarios. 
By identifying the strengths and limitations of both methods, our paper contributes to the development of a reliable AI-based assistant to process VAT-related queries.
The source code of the implementation, the datasets for the experiments, and the results are available in an online repository~\cite{benkel_2026}.

With respect to the current published literature, this paper has the following contributions:

\begin{itemize}
    \item The research contributes to a more systematic assessment of the legal reasoning abilities of large language models (LLMs) in practice, beyond anecdotal evidence and general-purpose benchmarks on generic synthetic data. 
    The research presents an experimental evaluation on both textbook cases from authoritative sources and complex real-world cases from tax consulting practice.
    
    \item The research empirically compares fine-tuning and retrieval-augmented generation (RAG) in tax law on textbook and real-world cases, offering valuable insights into their suitability for domain specialization of LLMs in Austrian and EU VAT law.
    
    \item The research specifically highlights the strengths and limitations of both approaches---fine-tuning and RAG---in terms of accuracy, explainability, and adaptability.
    The insights derived from applying these methods to complex real-world VAT cases support the advancement of AI-assisted legal advisory systems in tax consulting.
\end{itemize}

In this paper, we focus on improving the reasoning capabilities of large language models (LLMs) in the context of VAT law, primarily by applying fine-tuning and RAG. 
Fine-tuning and RAG directly address well-known limitations of foundation models, such as outdated information, insufficient domain-specific knowledge, and hallucinations~\cite{ibrahim2025}. 
Given that our work addresses the highly sensitive domain of VAT law, where incorrect or hallucinated information may lead to financial penalties, regulatory disputes, or loss of trust among tax professionals, our methodological priority was to reduce hallucinations and ensure verifiable, legally grounded answers. 
For these reasons, we deliberately focused on fine-tuning and RAG as techniques that strengthen domain knowledge and factual grounding.

The remainder of this paper is organized as follows.
In Section~\ref{sec:background}, we provide background information and review related work.
In Section~\ref{sec:methodology}, we describe the methodology.
In Section~\ref{sec:cases}, we present the datasets used for experimental evaluation.
In Section~\ref{sec:prompt_engineering}, we discuss prompt engineering.
In Section~\ref{sec:rag}, we present the approach for RAG.
In Section~\ref{sec:finetuning}, we present the approach for fine-tuning LLMs for decision-making in VAT cases.
In Section~\ref{sec:evaluation}, we conduct a final evaluation of (variants of) the developed AI-based assistant.
In Section~\ref{sec:discussion}, we discuss the results.
In Section~\ref{sec:conclusion}, we conclude with a summary and an outlook on future work.

\section{Background and Related Work}\label{sec:background}

In the following, we review related work and provide background information on VAT law.

\subsection{Large Language Models in Legal Decision-Making}

LLMs demonstrate varying levels of accuracy in legal decision-making, ranging from 19.2~\% to 97.6~\% in empirical studies (see~\cite{ElHamdani2024TheFO,nay2024,Jayakumar2023LargeLM,fei_lawbench_2023,savelka2023unreasonable,savelka2023unlocking,ammar2024prediction}), with performance dependent on the specific legal domain, task type, and whether the model is specifically trained for legal applications. 
These empirical studies assess LLMs' accuracy and reliability in legal decision-making across varied domains, including Chinese civil, contract, and criminal law, US fiduciary and tax law, Arabic commercial law, and case law and legislation. 
In classification, extraction, and generation tasks, one study~\cite{fei_lawbench_2023} reports that GPT‑4 scores average 52--54~\% (with some tasks reaching 97.6~\%).
In legal judgment prediction for Chinese criminal law~\cite{fei_lawbench_2023} GPT‑4 reaches 74.5~\% accuracy, while another study~\cite{nay2024} on fiduciary obligations finds accuracy ranging from 27~\% (using Curie) to 81~\% (using GPT‑3.5). 
Models that are fine‑tuned on legal data, such as SaulLM and LegalBERT, show improved performance on tasks closely aligned with their training, whereas general LLMs (GPT‑4, GPT‑3.5) typically generalize better in few‑shot or zero‑shot settings~\cite{ElHamdani2024TheFO}.
Performance depends strongly on domain (e.g., Chinese civil, contract, criminal law, United States fiduciary and tax law, Arabic commercial law), task type (classification, extraction, generation), and legal-specific training, meaning that research on Austrian tax law.

Research implies that there is a need for further development to improve LLM performance in legal contexts, that legal-specific fine-tuning is beneficial, but that larger models and better evaluation metrics are required~\cite{fei_lawbench_2023}.
Furthermore, this research shows that accuracy of LLMs was examined through various legal sources. 
Our paper expands this research for a specific field of tax law, the Austrian and EU VAT law.
Comparable studies examining the capabilities and accuracy of LLMs dealing with Austrian and EU VAT cases have not been conducted yet.
Thus, this research adds a new idea and a new perspective to the existing knowledge in the field of application of LLMs for legal purposes.

Recent research on LLMs in tax law has focused on the use of LLMs in tax administration and the capabilities of LLMs in applying tax law, aiming to understand their potential in legal analysis. 
Kuźniacki et al.~\cite{gorski2024exploring} explore the use of AI by tax administrations for automation purposes and its potential impact on the rights of taxpayers.
Kuźniacki et al.~\cite{gorski2024exploring} argue that adequate protection of taxpayers' rights demands the use of explainable AI technologies that can render the functioning and decisions of tax AI systems understandable for persons affected by automation of tax law.
Kuźniacki et al.~\cite{gorski2024exploring} also found that LLMs demonstrate emerging legal understanding, with performance improving across successive OpenAI model releases. 
The study revealed that providing relevant legal context and using few-shot prompting significantly enhanced the performance of advanced models like GPT-4~\cite{nay2024}. 
Sun~\cite{sun2023shortsurveyviewinglarge} points out that LLMs are increasingly being applied in the legal domain, including for tasks like legal judgment prediction and document analysis. 
However, according to Sun~\cite{sun2023shortsurveyviewinglarge}, the integration of LLMs into legal practice poses challenges such as privacy concerns and bias. 
Yue et al.~\cite{yue_disc-lawllm_2023} propose DISC-LawLLM, an intelligent legal system utilizing LLMs to provide different legal services, and DISC-Law-Eval, a benchmark to evaluate the LLM in both objective and subjective dimensions. 
The results demonstrate the effectiveness of DISC-LawLLM in serving various users across diverse legal scenarios.
Furthermore, a Chinese LLM~\cite{xu2024surpassing} surpassed human performance in professional tax qualification exams, demonstrating the potential of LLMs in specialized domains.
This study applied a unique training methodology involving multitask and single-task model fine-tuning which was found to be more effective than traditional methods.

Regarding the performance of the two predominant approaches for improving the factual accuracy and general quality of an LLM's outputs described in the literature, namely, RAG and fine-tuning, a clear and consistent evaluation  in the specific legal domain of tax law is missing. 
A literature review reveals that various related works have have enriched LLMs with specific legal knowledge via RAG or fine-tuned LLMs for legal decision-making. 
But what about systematically comparing the performance of both approaches based on real-world cases? 
Literature does not address the question whether both approaches are necessarily required in the legal domain or whether, for example, RAG alone could be sufficient for certain legal domains. 
Moreover, the literature states that LLMs demonstrate emerging legal understanding. 
But can we really talk about ``legal understanding''? 
Do LLMs posses ``legal understanding'' to reason about legal cases or is it more about solving cases based on juridical methods which follow certain logical rules? 
And what about the ``legal understanding'' of LLMs and the performance of RAG and/or fine-tuning based on real-world cases resulting from tax advisory?
Our paper deepens the research on fine-tuning and RAG in tax law, as it measures and compares the results of both approaches using real-world cases, providing valuable insights into the suitability of fine-tuning and RAG for the domain specialization of LLMs in Austrian and EU VAT law. 
Our paper investigates the advantages and disadvantages of both approaches in terms of the factual correctness of the decisions and the provided justifications; correct justifications would add explainability to the decisions.
Furthermore, the insights gained from the application of RAG and fine-tuning to real-world VAT cases contribute to the further development of AI-assisted legal advisory systems in tax consulting. 
Moreover, this paper contributes a further component to the research field of legal reasoning based on LLMs and RAG. 
Our paper relates to the work conducted by Fei et al.~\cite{fei_lawbench_2023} and Yue et al.~\cite{yue_disc-lawllm_2023}, providing an approach for more usable and reliable LLMs and thus making them applicable across more legal domains. 
Finally, our paper complements the findings of Wiratunga et al.~\cite{wiratunga2024cbr}, Nay et al.~\cite{nay2024} and Fei et al.~\cite{fei_lawbench_2023} delivering specific insights into the reasoning capabilities of LLMs for VAT purposes and related accuracy based on different configurations. 
Eventually, by combining systematic, automated evaluation with targeted improvements to both RAG and fine-tuning, our paper contributes to the development of a robust and efficient AI-based legal reasoning system for VAT law.

A consistent finding across the literature is that domain-specific or adapted models significantly outperform general-purpose LLMs on legal tasks. 
For example, Hou et al.~\cite{hou2025} identify 16 legal LLM model series and 47 LLM-based frameworks tailored to legal applications. 
Similarly, Fei et al.~\cite{fei_lawbench_2023} evaluate 51 LLMs across 20 legal tasks, showing that while general-purpose models such as GPT-4 perform strongly, they still lag behind specialized legal models. 
This performance gap is further quantified by Jayakumar et al.~\cite{Jayakumar2023LargeLM}, who demonstrate that general-purpose LLMs (e.g., ChatGPT-3.5, LLaMA-70B, Falcon-180B) underperform compared to smaller, fine-tuned models for a particular domain by 19.2--26.8~\% on contract classification tasks.
Existing applications of LLMs in law include legal judgment prediction, case retrieval, and contract analysis~\cite{qin2024}. 
At the same time, key challenges persist, including hallucinations, limited explainability, bias, and ethical concerns~\cite{shao2025}. 
Domain-adaptive pre-training has emerged as a promising direction, with models such as Phi-2-Legal and Mistral-Legal-7B~\cite{arbel2024} achieving strong performance through continued pre-training on large-scale legal corpora.

Han et al.~\cite{han2025} investigate a substantially different research question than our study, evaluating LLM approaches for legal-citation prediction, specifically focusing on improving the accuracy with which models identify and generate correct legal references within the Australian legal system. 
The central challenge addressed in that study is that LLMs frequently produce incorrect or fabricated citations.
Han et al. compare multiple techniques to alleviate that problem, including domain-specific pre-training, fine-tuning, RAG, and ensemble methods, with the goal to determine which approach most accurately predicts relevant statutes or precedents. 
Although the paper refers to an ``Australian Law Case study,'' the legal domain primarily serves as the context for benchmarking of citation prediction rather than for the substantive analysis or reasoning over legal cases themselves.
In contrast, our study investigates how LLMs can reason about and explain VAT law, focusing on generating legally sound interpretations and explanations rather than predicting citations. 
Consequently, the methodological emphasis and evaluation criteria differ substantially from Han et al.~\cite{han2025}, which explains why our work concentrates specifically on fine-tuning and RAG as mechanisms for improving reasoning and factual reliability in tax law analysis.

Recent studies explore the use of external knowledge to improve LLM performance.
For example, Doyle et al.~\cite{doyle2025} investigate how injecting excerpts from legal practice guides into prompts affects LLM reasoning. 
The results show that retrieval-style augmentation can improve performance in tasks such as legal question answering and case outcome prediction across multiple U.S. jurisdictions.
Similarly, LegalAgentBench~\cite{li2025} evaluates LLM-based agents in complex legal workflows, highlighting the importance of tool use and structured interaction with external information sources.
However, these studies primarily evaluate whether augmenting LLMs with external knowledge improves performance, rather than systematically comparing different adaptation strategies. In particular, they do not provide a controlled, head-to-head comparison between retrieval-augmented generation (RAG) and fine-tuning approaches.

Prior surveys and benchmark studies further confirm the rapid growth of the field of legal LLMs. 
For example, Dheghani et al.~\cite{dehghani2025} identify multiple LLM architectures and frameworks for the legal domain, while \emph{LegalBench}~\cite{guha2023} represents an effort to compare various LLMs across a wide range of legal tasks. 
In this regard, existing studies highlight both the potential and limitations of general-purpose LLMs in legal settings, particularly with regard to hallucinations, reasoning reliability, and the need for domain adaptation.
However, most prior work focuses either on benchmarking LLM performance on legal tasks, evaluating LLM-based agents, or developing domain-specific legal models, rather than directly comparing methods for improving reasoning quality in legally sensitive decision-making contexts. 
In contrast, our study focuses specifically on VAT law reasoning using real-world tax cases, where factual accuracy and legally grounded explanations are critical. 
In addition, while previous work often explores domain-adaptive pre-training or benchmark evaluations, our research directly compares fine-tuning and retrieval-augmented generation (RAG) as two practical approaches for improving the reasoning performance of LLMs in a specialized legal domain. 
This head-to-head comparison addresses an important gap in the literature, as relatively few studies systematically evaluate these two techniques in the context of real legal case analysis.

In summary, the literature reveals a clear research gap. 
While prior work extensively studies (i) domain-specific fine-tuning and (ii) retrieval-augmented approaches in isolation, there is little systematic, controlled comparison between these two paradigms in legal applications.

As a form of fine-tuning, \emph{instruction tuning} serves to teach LLMs to follow specific types of instructions (see~\cite{bergmann2026} for further information).
Our fine-tuning approach could also be understood as a form of instruction tuning, i.e., a specialized type of supervised fine-tuning in which a pretrained language model is trained on task-oriented input-output pairs.
In our case, the input consists of textual descriptions of tax scenarios, while the target outputs are corresponding expert analyses. 
Although the instructions are implicit rather than explicitly phrased as prompts (e.g., ``analyze this case''), the training setup still teaches the model to map problem descriptions to structured, domain-appropriate responses, thereby learning to perform the underlying task. 
Consequently, our method aligns with instruction tuning in that it focuses not only on adapting the model to the tax domain, but also on shaping the model's behavior to generate coherent and legally-founded analyses in response to case descriptions.

Prompt design has also been shown to significantly influence LLM performance in the legal domain, with effects varying depending on task complexity and domain specificity.
Related work~\cite{hakimi_parizi2023} shows that different prompting strategies (e.g., zero-shot, few-shot, chain-of-thought, and prompt ensembling) can lead to substantially different outcomes, and that not all strategies transfer equally well to legal tasks due to long documents and specialized language.
Other work~\cite{zhang2024} further shows that using empathetic language in the prompts can improve output quality, although domain knowledge remains a critical factor for achieving strong performance.
More recent work~\cite{sriram2026} finds that context-layered prompts and chain-of-thought reasoning, yield the largest improvements, particularly for complex legal reasoning tasks, while also reducing hallucination rates.
These findings suggest that the effectiveness of prompting is highly task-dependent and interacts with both model capabilities and the availability of additional background knowledge, e.g., in the form of legal documents.
In the context of our work, we control for prompt design to ensure a fair comparison between fine-tuning and RAG.
Specifically, we apply consistent prompting strategies across all experimental conditions, allowing us to attribute performance differences primarily to the underlying adaptation approach rather than prompt variation.
While alternative prompting strategies could further improve absolute performance, our results focus on the relative comparison between methods under controlled conditions.

\subsection{Value-Added Tax Law}

In the following, we briefly explain the legal basis regarding the concept of VAT in Austria and the influence of EU provisions on Austrian VAT law. 
In Austria, VAT, also known as ``tax on turnover'', is a consumption tax applied to most goods and services. It is based on the value added at each stage of production and distribution. From a federal revenue point of view, VAT constitutes, besides wage tax, the major source of income for the Austrian Federal Government~\cite{statistics_austria_tax_2024}. This means that VAT revenue is essential for public finances, but only the state where the supply of goods and services takes place, according to VAT rules, is entitled to collect it.
According to Section~1 of Austrian VAT law~\cite{ustg_1994_2024}, taxation is based on the place of supply: VAT is levied only if the supply of goods or services occurs within Austria. If the place of supply is abroad, Austria cannot impose VAT; instead, the respective foreign jurisdiction has the right to tax the transaction.

The \emph{place of supply of goods} is determined by the nature of the goods supplied and on the mode of supply. According to Section 3 of Austrian VAT law~\cite{ustg_1994_2024}, a supply of goods is deemed to have taken place within Austria if the goods were located in Austria at the point at which the power of disposition was transferred. However, in the case of dispatch or transport of goods, the place of supply is the place where the goods are located when the dispatch or transport to the customer begins.
The following examples, which are similar to those in the dataset used for experiments (see Section~\ref{sec:textbook_cases}), illustrate the basic principles of determination of the place of supply of goods. For example, if a Croatian manufacturer is hiring a machine in Croatia from an Austrian company and decides to purchase the hired machine, the place of supply is Croatia. Consequently, Croatian VAT law applies and Croatia has the right to levy VAT on the supply of the machine. The reason for this is the fact that the power of disposition was transferred to the purchaser in Croatia. Notwithstanding, in the case the machine supplied is subsequently transported from Vienna to a customer in Zagreb (Croatia), the supply would be subject to Austrian VAT because the transportation started in Vienna (Austria). It is important to mention that these basic rules for the determination of the place of supply are in line with EU VAT provisions, since Austrian VAT law is harmonized by EU law. However, depending on the nature of the transaction at hand, additionally other special EU VAT provisions may apply, e.g., in case of chain transactions or triangular cases. This leads to more complex legal cases and requires careful and more transparent legal reasoning.

The \emph{place of service} is determined by Section 3a of Austrian VAT law \cite{ustg_1994_2024} containing an extensive list of rules to determine the place at which a service is carried out. This depends on the nature of the service but also on the status of the person receiving the service. This means it must be differentiated between taxable persons (business-to-business) and non-taxable persons (business-to-customer). In the case of services provided to taxable persons, the place of supply is generally where the recipient is established (business-to-business general rule). For non-taxable persons, it is where the service provider has established their business (business-to-consumer general rule). As with the place of supply of goods, EU provisions affect also the place of service provision. Thus, besides the Austrian VAT rules, also the EU VAT provisions must be taken into consideration and applied where necessary.
As another example for determining which country is entitled to levy VAT, consider accountancy services provided by a Bulgarian company to a business customer with place of business in Austria. In this case the place of service provision is located in Austria according to Section 3a of Austrian VAT law and EU VAT provisions. Or, in case of advertising services rendered by an Austrian company to an enterprise situated in Italy, the place of provision of the advertising service is Italy, because the company receiving this service is situated in Italy. Contrary to this, in case of a consultancy service company established in Austria providing services to a customer (individual) who resides in Denmark, the place of service rendering would be located in Austria. 
As already mentioned, numerous exceptions to the general rules exist, making the determination of the place of supply complex. The illustrated examples offer only a brief overview of this topic.

Austrian VAT law was enacted by the VAT Act of August 23, 1994, effective from January 1, 1995, and amended by Federal Law (BGBl I 201/2023) in 2023. The VAT Act implemented the EU single market regulations~\cite{european_union_directive_2006}, ensuring harmonization with the broader EU VAT system. Consequently, the Austrian VAT law consists of two parts: one governing domestic and non-EU transactions, and the other addressing intra-EU transactions. The latter are known as the EU single market regulations, which constitute an appendix to Section 29 of Austrian VAT law. Besides the legal sources, the Directive of the Austrian Federal Ministry of Finance~\cite{umsatzsteuerrichtlinien_2000_ustr_2023} provides interpretative guidance and practical examples for applying VAT law to individual cases.

\section{Methodology}\label{sec:methodology}

We follow Wieringa's framework~\cite{wieringa2014} for design science research (DSR) in information systems.
Wieringa distinguishes between \emph{knowledge questions} and \emph{design problems}.
Whereas knowledge questions ``ask for knowledge about the world'', design problems ``call for a change of the world''~\cite[p.~6]{wieringa2014}.
The formulation of knowledge questions often leads to the formulation of design problems, with knowledge questions and design problems being iteratively revisited.

Previous research has shown that LLMs already possess considerable reasoning abilities in the legal domain~\cite{martin_better_2024}, especially in tax law~\cite{nay2024}.
Anecdotal evidence from experts in tax management and tax technology confirms the findings reported in the literature.
However, we are interested in a more systematic, empirical analysis of LLMs' capabilities and pick an area of tax law---VAT law---that is internationally harmonized at EU and OECD levels, requires a variety of decisions for correct accounting, and occurs regularly in daily operations, making the results of such an analysis more generalizable.
We are also particularly interested in the applicability of two common approaches to enhancing an LLM's reasoning capabilities: fine-tuning of an LLM to deciding cases in VAT law and the use of RAG to provide LLMs with required background information on tax law.
Thus, we start with the following knowledge questions.

\begin{itemize}
  \item[\textbf{RQ1}] \textit{What capabilities do LLMs have to reason about real-world cases in Austrian VAT law when enhanced with fine-tuning or RAG?}
  
  \item[\textbf{RQ2}]\textit{Which approach---fine-tuning or RAG---better enhances an LLM's capability to provide legally-justified decisions on cases in Austrian VAT law?}
\end{itemize}

To answer the identified knowledge questions, we develop (variants of) an AI-based assistant that takes a natural-language description of a VAT case and returns a legally justified decision on the VAT case.
In daily tax consulting practice, clients often report cases using natural-language descriptions, for example, in an email.
A tax consultant must interpret these descriptions and return a legally justified decision on the case.
An AI-based assistant for automating decisions would greatly improve the efficiency of decision-making in VAT cases.
Thus, the knowledge questions lead to the following design problem.

\begin{itemize}
    \item Improve the efficiency of decision-making in VAT cases described in natural language (\emph{problem context})
    \item by designing an AI-based assistant (\emph{artifact})
    \item that automatically provides accurate decisions on VAT cases and correct justifications for the decisions (\emph{requirements})
    \item so that VAT consultants are relieved of routine tasks (\emph{stakeholder goals}).
\end{itemize}

\begin{figure}[ht]
    \centering
    \includegraphics[width=0.65\textwidth]{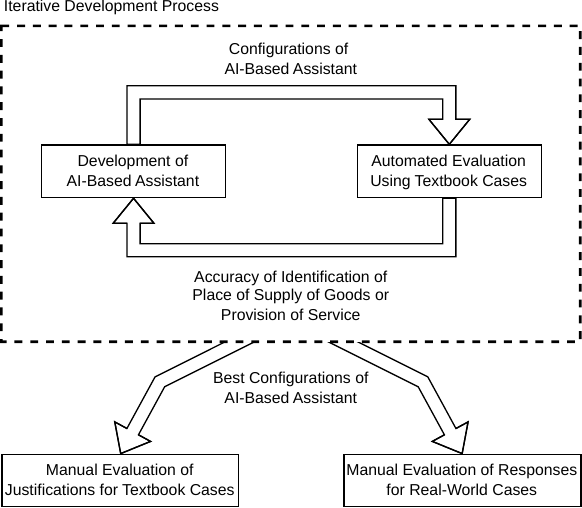}
    \caption{Iterative development process of the AI-based assistant and final evaluation}
    \label{fig:methodology-iterative_development}
\end{figure}

We iteratively develop variants of an AI-based assistant with an LLM as the backbone, using different configurations (Figure~\ref{fig:methodology-iterative_development}).
The development iterations serve to optimize the configurations, in particular prompt messages as well as various parameters for RAG and fine-tuning.
In each iteration, the different configurations are automatically evaluated using textbook cases from Berger and Wakounig~\cite{berger_umsatzsteuer_2022} about the place of supply of a good or provision of a service, the identification of which is required in international transactions to determine the country where the tax shall be levied.
The accuracy of the AI-based assistant in identifying the place of supply is a simple metric that can be easily calculated and used for rapid development of various iterations of the AI-based assistant.
The textbook cases thus serve as a \emph{validation set} during the development.

We perform a final evaluation of the best configurations found in previous development iterations using both textbook cases and real-world cases.
In this final evaluation, we first manually evaluate the reasoned justifications of the identified place of supply of goods or provision of service for the textbook cases; these justifications provide \emph{explainability} of the decisions.
We then manually evaluate the responses for real-world cases from the daily practice of a tax consulting firm, which go beyond the mere determination of the place of supply of goods or provision of service, thereby investigating the \emph{adaptability} of LLMs for different reasoning tasks in VAT law.
In the course of this manual evaluation, we also consulted additional tax experts for their opinions regarding the correctness of the system's responses.
The real-world cases thus serve as a \emph{test set} for evaluating the previously identified best configurations of the AI-based assistant.

The performance of the various configurations of the AI-based assistant is measured in terms of accuracy.
During iterative development, the term ``accuracy'' refers to the percentage of cases in which the predicted country (place of supply) matches the ground truth.
This evaluation can be automatically conducted and allows for fast iterative development.

Following the automated evaluation of the correctness of the identified place of supply in the textbook cases, the justifications for the textbook cases provided by the best configurations of the AI-based assistant are subjected to a secondary validation phase involving human tax experts. 
The justifications provided for answers that were classified as correct during the automated evaluation---based on the place of supply of goods or provision of services at the country level---are reviewed by three human tax experts, two of which were otherwise not involved in the research project to ensure an unbiased assessment.
Thus, the justifications provided by the LLM for determining the applicable country were classified as either \emph{correct} or \emph{incorrect}.

The evaluation of the AI-generated justifications was a collaborative effort between the human experts, with the experts conducting a mutual cross-check of their evaluations, thereby minimizing potential errors in the assessment process, ultimately reaching a consensus.
Computing initial inter-rater agreement is of limited relevance in this setting, as the evaluation objective is to produce a single, consensus-based evaluation rather than to measure independent judgments. 
The objective of the expert assessment is not to quantify variability between raters, but to resolve the variability through deliberation in order to establish a legally coherent evaluation of the AI-generated justifications.

The manual assessment of justifications is critical, as a response could identify the correct country but provide an incorrect legal justification (e.g., citing incorrect legal provisions).
Since the justifications are classified into \emph{correct} and \emph{incorrect}, the accuracy of the AI-based assistant in returning justifications is the percentage of justifications for cases that are correct out of all the textbook cases.

The human tax experts also manually classify the analyses for the real-world VAT cases as either \emph{correct} or \emph{incorrect}, likewise achieving consensus through mutual cross-checking of the evaluations.
This manual assessment considers multiple factors, including the substantive quality of the analysis, the correctness of the cited legal provisions, and the appropriate use of relevant tax terminology.
The accuracy of the configurations of the AI-based assistant in returning correct analyses of real-world VAT cases is the percentage of analyses that are correct out of all real-world VAT cases.

Unlike classification tasks in machine learning, \emph{precision} and \emph{recall} cannot be meaningfully defined in the context of this study.
The task at hand is not a classification task.
For classification tasks, precision is the percentage of relevant instances of a class that were retrieved by a classifier out of all retrieved instances, while recall is the percentage of relevant instances of a class that were retrieved by a classifier out of all relevant instances.
In the context of this study, we are interested in the tasks of identifying the place of supply in a VAT case, providing the corresponding justifications, and providing a general analysis of more complex real-world cases.
We measure the correctness of these tasks in terms of accuracy.
The accuracy values for the different tasks provide a comprehensive assessment of the ability of LLMs to reason about VAT cases in tax law.

We conduct technology-oriented experiments to compare the different configurations of the AI-based assistant.
According to Wohlin et al.~\cite[p.~73]{Wohlin2024}, an experiment in software engineering is ``an empirical enquiry that manipulates one factor or variable of the studied setting''.
In experiments, ``different treatments are applied to or by different subjects, while keeping other variables constant, and measuring the effects on the outcome variables''~\cite[p.~73]{Wohlin2024}.
In contrast to human-oriented experiments, in technology-oriented experiments, ``different technical treatments are applied to objects''~\cite[p.~73]{Wohlin2024}, e.g., by applying different tools to the same objects~\cite[p.~74]{Wohlin2024}.
In the studied setting, the different technical treatments (or tools) are the various configurations of the AI-based assistant, which are applied to VAT cases, and the outcome variable is the correctness of the case decisions returned by the AI-based assistant.
We manipulate various configuration parameters to study the effect on the accuracy of the AI-based assistant's outputs to gain an understanding of LLMs' reasoning ability in the area of VAT law.

To isolate the effect of different optimization approaches, namely, RAG versus fine-tuning, we conduct the experiments using models from a single LLM family provided by OpenAI, which we chose based on the results of \emph{LegalBench}~\cite{guha2023}, where the GPT models outperformed other LLMs in legal reasoning tasks. 
Holding the base model family constant allowed us to systematically vary prompting, retrieval, and fine-tuning configurations while minimizing confounding factors arising from architectural differences between models. 
This research design ensures that observed performance differences can be attributed to the optimization approach rather than to underlying model variations.

The widening of LLMs' context windows means that all legal sources related to VAT can now be included directly into the prompt without the need for RAG.
Nevertheless, including more documents in the prompt increases the cost of using LLMs compared to a RAG-based configuration.
Therefore, if a RAG-based configuration achieves performance comparable to that of a long-context LLM with all relevant legal sources included in the prompt, the RAG-based configuration would offer a more favorable cost–benefit ratio.
Furthermore, many documents or chunks of documents are not relevant for answering a question.
Including irrelevant chunks in the prompt could potentially lead the LLM into the wrong direction.
Thus, we also conduct experiments with GPT-4.1, which has a context window of 1~million tokens~\cite{openai_gpt41_2025}, to compare the performance of a long-context LLM with the legal sources embedded in the prompt to the performance of the configurations with RAG and fine-tuned LLM.

\section{VAT Cases for Experimental Evaluation}\label{sec:cases}

In the following, we briefly describe the cases used for the experiments.
We first describe the textbook cases that were used for continuous validation of different configurations during the development cycles.
We then describe the real-world cases used to test the best configurations of the AI-based assistant found during development.

\subsection{Textbook Cases}\label{sec:textbook_cases}

For evaluation during development, we used 74 selected cases from the textbook \textit{Umsatzsteuer kompakt 2022/2023} (in English:
\textit{Value-Added Tax in a Nutshell 2022/2023}) by Berger and Wakounig~\cite{berger_umsatzsteuer_2022} as a validation dataset. 
The selected cases focus on determining the place of supply of goods or the provision of a service according to Austrian VAT law. 
For each case, the correct place of supply at the country level and the legal justification were also manually determined to serve as the ground truth for the performance evaluation.
In addition, the dataset contains the chunks from the legal sources relevant to solving the case, which serve to evaluate the the RAG system's retrieval performance.

The individual cases that make up the dataset are collected in tabular form, along with the correct place of supply or service provision, the correct justification, and the applicable legal texts.
The dataset is converted into JSONL format for further processing.
Both the tabular and JSONL versions of the dataset consisting of textbook VAT cases are available online~\cite{benkel_2026}.

The size and composition of the dataset follow from the study's objective of systematically covering a well-defined legal domain rather than approximating a large, randomly sampled population. 
The 74 textbook cases were selected to reflect the full range of constellations arising from the statutory rules governing the determination of the place of supply under Austrian VAT law.

The textbook cases, while inspired by real-world practice, are fictitious.
For didactic reasons, textbook cases are often idealized descriptions to illustrate various principles of tax law.
In general, textbook cases are easier to solve than cases encountered in real-world practice of tax consulting.
Thus, we also evaluate LLM performance on cases from real-world practice.

\subsection{Real-World Cases}

The dataset used to evaluate the AI system's performance in real-world VAT cases was provided by the tax consultancy firm \emph{ICON Wirtschaftstreuhand GmbH}.
The dataset includes 20 client inquiries that span a wide range of VAT-related issues, going well beyond questions related to the place of supply and service provision. 
Each inquiry references multiple sections and provisions of the Austrian VAT Act, ensuring a comprehensive and diverse scope. 
This variety allows for a robust evaluation, enabling a thorough assessment of the model's ability to handle the full range of complex, real-world VAT scenarios.

The real-world VAT cases were originally provided in the form of email correspondences, including client inquiries and responses drafted by professional tax advisors. 
These responses contained detailed justifications and explanations based on applicable VAT regulations, which serve as the basis for later evaluation of the responses given by the AI-based assistant.
To ensure compliance with data protection and confidentiality standards, all client-specific information was anonymized as a first step. 
Following anonymization, the cases were systematically organized into a tabular format to structure the key elements of each inquiry and response, thereby creating a coherent and analyzable dataset.

In some cases, the email correspondences referenced attachments---for example, contracts or other supporting documents---that were not included in the dataset.
Because such materials often form a crucial basis for tax assessments, the advisors' responses occasionally contained implicit information inferred from these missing sources, which the AI-based assistant would have no possibility to infer from the case description.
To partially compensate for the absence of the original attachments, relevant details were extracted from the advisors' explanations and added to the corresponding inquiry, thereby enriching the input available to the model without revealing the correct answer.
Although this approach aimed to approximate the informational value of the missing documents as closely as possible, it cannot be ruled out that certain context-specific details remained inaccessible due to the absence of the original attachments.
The dataset with the real-world VAT cases is available online~\cite{benkel_2026}.

\section{Prompt Engineering}\label{sec:prompt_engineering}

We develop different prompts for different purposes.
We employ a special prompt for the specific question regarding the place of supply or service provision in a VAT case, which is essential for establishing the country entitled to levy VAT in international transactions.
We employ another prompt for the general legal analysis of VAT cases.
In the following, we briefly discuss the prompts and the development process that led to their final formulation.

\subsection{Identification of Place of Supply}

To determine the best prompt formulation, we used the textbook VAT cases on the question of the place of supply.
We evaluated the quality of the prompts based on the accuracy of the identified places of supply, using plain GPT-4 and GPT-4o models as well as preliminary versions of the fine-tuned variants of these models;
the fine-tuning was performed on a subset of the fine-tuning dataset (see Section~\ref{sec:finetuning}).

We started with an instructive prompt formulation that follows OpenAI's prompt engineering guidelines~\cite{openai_prompt_2024}.
We developed three additional versions of the prompt message with reduced instructions compared to the first version.
These prompt messages had more precise and less extensive instructions, which improved the accuracy in the textbook VAT cases.
The final prompt message then emphasizes that there can only be one place of taxation, addressing the previously recurring issue of models outputting multiple countries. 
References to the tax regulations of the Austrian VAT Act are required to be included in the justification. 
The responses must be in German.
The output of the results in JSON format allowed the country and justification to be separated with Python code to automatically evaluate correctness. 
Arranging the JSON keys so that the justification precedes the place of supply or service provision resolved a recurring issue observed in previous prompts where the output of the country sometimes contradicted the justification.

\begin{lstlisting}[
float=t, 
caption=Template for prompt message used in the experiments for identification of place of supply or service provision,
label=lst:prompt_basic,
basicstyle=\ttfamily\footnotesize, 
frame=single, 
breaklines=true, 
breakatwhitespace=true,
numbers=left
]
system:
You are a tax assistant who is supposed to determine the place of taxation for each question. This can only be one place. 
Always answer in German.
Justify the place of supply according to the Austrian Tax Act.
Analyze the question carefully and decide which (special) regulation applies.

Always return a JSON with the following keys:
* justification: A well-founded justification for the place of supply based on the applicable paragraphs or articles of the Tax Act.
* country: The country that represents the place of supply (place where tax is paid).

context: {{contexts}}
user: {{question}} 
assistant:
\end{lstlisting}

Listing~\ref{lst:prompt_basic} shows the finally adopted template for the prompt message to identify the place of supply of goods or provision of a service.
RAG systems insert additional background information, i.e., chunks of legal sources, in place of \texttt{\{\{contexts\}\}} in the prompt message.
For system configurations without RAG, no additional background information is included as context, i.e., Line~11 is removed.
The user question replaces \texttt{\{\{question\}\}} in the prompt message submitted to the LLM.

\subsection{General Analysis of VAT Cases}

The prompt message in Listing~\ref{lst:prompt_basic} was the basis for the development of a prompt message for general analysis of the real-world VAT cases, which goes beyond the comparatively simple identification of the place of supply.
From this prompt message, the elements that specifically relate to the place of supply of goods or provision of service were removed.
The prompt message has been further modified to explicitly request a careful analysis of the VAT case. 

\begin{lstlisting}[
float=t, 
caption=Template for prompt message for the general analysis of real-world VAT cases, 
label=lst:prompt_general_rag,
basicstyle=\ttfamily\footnotesize, 
frame=single, 
breaklines=true, 
breakatwhitespace=true,
numbers=left
]
system:
You are a tax assistant who is supposed to answer tax-related questions from a legal perspective. 
Refer to the relevant paragraphs and articles of the Austrian VAT Act.
Analyze the question carefully and decide which (special) regulation applies.
Always answer in German.

context: {{contexts}} 
user: {{question}} 
assistant:
\end{lstlisting}

Listing~\ref{lst:prompt_general_rag} shows the finally adopted prompt message for the general analysis of real-world VAT cases.
RAG systems insert additional background information, i.e., chunks of legal sources, in place of \texttt{\{\{contexts\}\}} in the prompt message.
For system configurations without RAG, no additional material is included as context, i.e., Line~7 is removed.
The user question replaces \texttt{\{\{question\}\}} in the prompt message submitted to the LLM.

\section{Retrieval-Augmented Generation}\label{sec:rag}

In the following, we describe the architecture of the implemented RAG system, the data sources, the preprocessing of those data sources, and the implementation of the RAG system in Azure AI Foundry. 
The implementation of the RAG system, the preprocessed data sources, and the results of the experiments are available online~\cite{benkel_2026}.

\subsection{Architecture}

General-purpose pre-trained LLMs that were not fine-tuned for decision-making in VAT cases may lack the contextual information required to correctly interpret a case. 
Even if the relevant information is present in the training data, the model may fail to retrieve or apply the information appropriately. 
For example, to decide VAT cases, the LLM must rely on the applicable legal acts and directives in their current versions, as in force in the relevant jurisdiction. 
To improve a general-purpose LLM's performance on specific tasks such as decision-making in VAT cases in Austrian law, external information can be explicitly incorporated into the answer-generation process.

\begin{figure}[t]
    \centering
    \includegraphics[width=\textwidth]{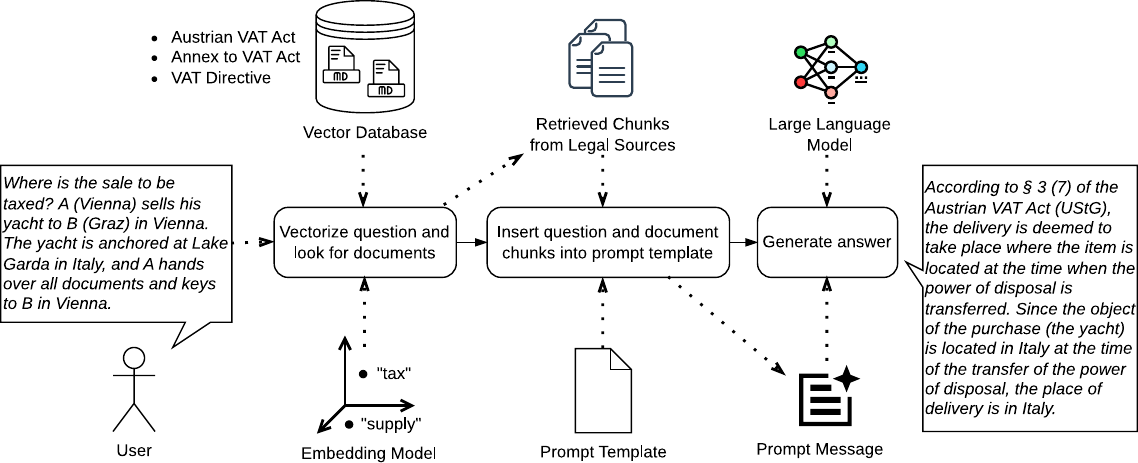}
    \caption{Process of retrieval-augmented generation}
    \label{fig:rag}
\end{figure}

RAG is a method to address the limitations of general-purpose LLMs when it comes to conducting specialized tasks; we refer to Gao et al.~\cite{gao_retrieval-augmented_2024} for more information.
Figure~\ref{fig:rag} illustrates the RAG process for decision-making in VAT cases, including its main components and the data artifacts involved.
The user begins by formulating a textual description of the case, which is submitted as input to the AI-based assistant.
The system then vectorizes the query using an embedding model and searches for document chunks with similar vector representations (embeddings) in a vector database containing relevant legal sources for VAT cases.
The retrieved chunks, along with the original user query, are inserted into a prompt template to form the complete prompt message.
This prompt message is then used with the LLM to generate the final answer.

\subsection{Data Sources}\label{sec:data_sources}
The RAG system has access to a collection of relevant legal documents, the purpose of which is to provide the contextual information required to correctly assess VAT cases. 
The legal documents reflect the current status of VAT law in Austria as of December 1, 2024.
These documents are official documents that are not subject to copyright restrictions.
In particular, the following documents, all in German, were selected.

\begin{itemize}
    \item Austrian VAT Act (74~pages)
    \item Annex to the Austrian VAT Act: Internal Market Regulation (18~pages)
    \item VAT Directive of the Austrian Federal Ministry of Finance (143~pages)
\end{itemize}

For decisions concerning the place of supply of goods or provision of services, only specific sections of the VAT Directive are required---namely, Sections~3 and~3a and Articles~1--4, which outline the relevant rules and provisions. 
In addition, the Austrian VAT Act and the Internal Market Regulation are consulted, resulting in a total of five documents required for such determinations.

In contrast, interpreting a broader range of real-world VAT cases requires access to the full text of the VAT Directive. 
The VAT Directive is segmented into 34 documents, each corresponding approximately to a specific section or article.
Combined with the Austrian VAT Act and the Internal Market Regulation, a total of 36 documents are to be consulted for decisions on real-world VAT cases.

Since the AI-based assistant should respond in German, we require that the documents be in German.
We note that the \emph{Annex to the Austrian VAT Act: Internal Market Regulation} is the Austrian implementation of the EU Internal Market Regulation.
EU directives and regulations are published in all official EU languages, including German.

In Austrian law, legal documents are typically structured into sections/articles (in German: \emph{Paragraph}, abbreviated: §), subsections (in German: \emph{Absatz}, abbreviated: \emph{Abs.}), items (in German: \emph{Ziffer}, abbreviated: \emph{Z.}), and letters (in German: \emph{Litera}, abbreviated: \emph{lit.}).
The different sections/articles vary widely with respect to the length as well as the number of subsections, items, and letters contained in the sections.
Some sections/articles comprise no further subsections at all, whereas others consist of numerous nested subsections, items, and letters.
For some cases, the entire article is required, whereas for others, only specific subsections, items, and letters are relevant.

The directives issued by the Austrian Federal Ministry of Finance (BMF) provide examples that have to follow the legal rules/norms and the jurisprudence of the Austrian Federal Finance Court. 
Both are, in general, legally binding for every person and institution underlying Austrian Tax Law. 
Consequently, the examples of the BMF directives have to follow the normative authority and contain examples that are solved following the legal rules and normative authority.
The BMF directives demonstrate the interpretation of the VAT law by the Austrian Ministry of Finance based on legal hierarchy and rules. 
In real-world tax consulting practice, the BMF directives are applied by Austrian tax advisers, to the extent that the BMF directives are a crucial part of the official exam for tax advisors.
When dealing with VAT cases and filing VAT returns, the financial authorities of the BMF are the institutions that have to assess the data and check if the cases are solved in accordance with applicable VAT law. 
Thus, the financial authorities are the first institutions issuing VAT notices, which shows the substantial relevance of the Austrian BMF directives when dealing with VAT cases.

\subsection{Data Preprocessing}

\begin{figure}[t]
    \centering
    \includegraphics[width=\textwidth]{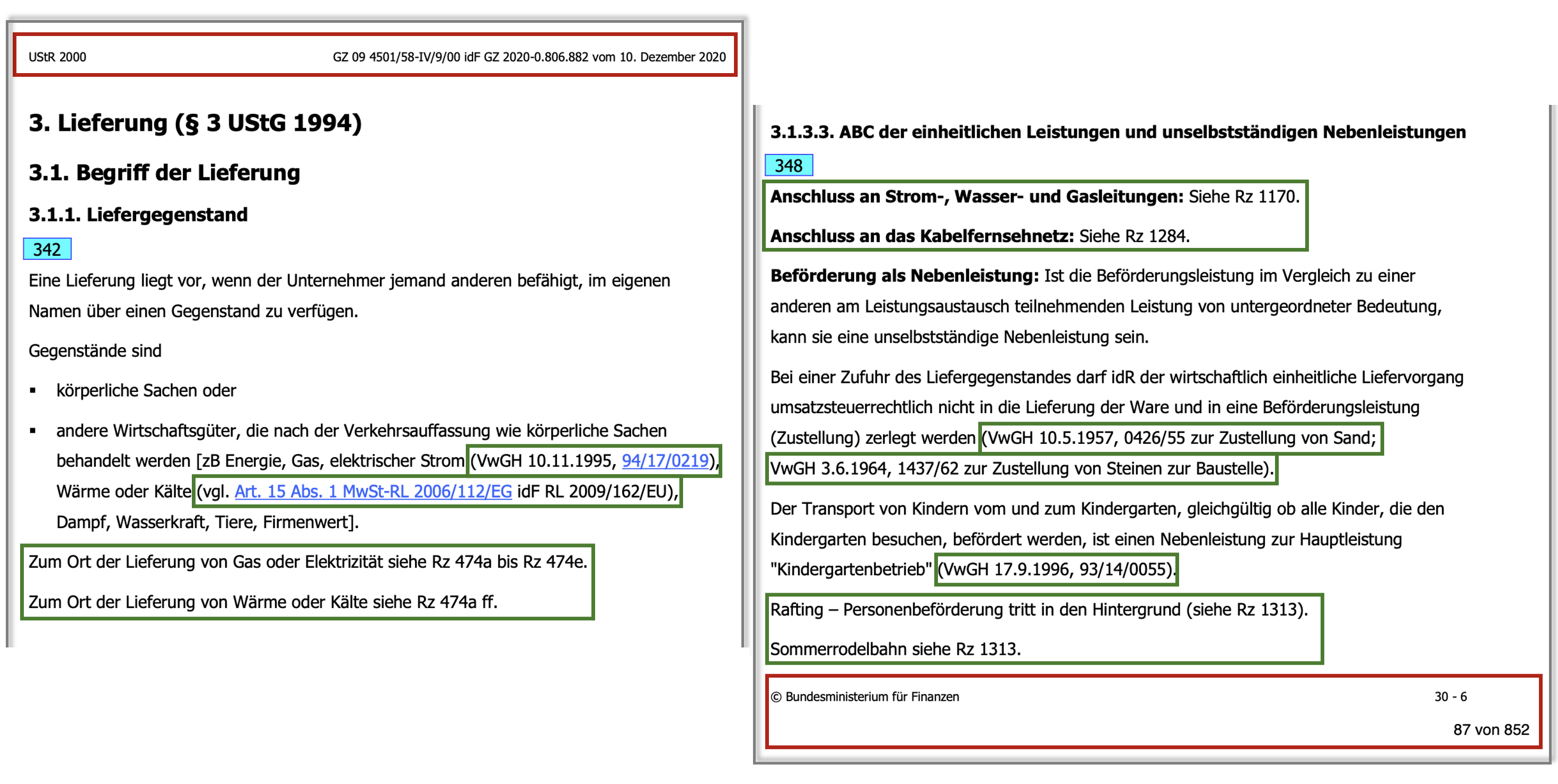}
    \caption{Preprocessing of legal documents in RAG.
    In the first step, header and footer content is removed.
    In the second step, the cross-references and other peripheral elements are excluded to ensure clean input for retrieval.}
    \label{fig:docs_preprocessing}
\end{figure}

To increase the effectiveness of the retrieval process, the PDF documents are first converted into Markdown format before removing extraneous header and footer content; see Figure~\ref{fig:docs_preprocessing} for an illustration of the removal of elements from the original data sources. 
This semi-automated process employed the \textit{PyMuPDF4LLM} library to extract text with Markdown formatting from the PDF files~\cite{noauthor_pymupdf4llm_2024}. 
Although the library provided a reliable baseline, manual refinements and regex-based adjustments were necessary to ensure accurate structure and formatting without altering the content.

We also experimented with the removal of cross-references and other peripheral elements, to obtain simpler and better-structured data.
We surmise that removing such extraneous elements, which is also illustrated in Figure~\ref{fig:docs_preprocessing}, will lead to more accurate justifications as well as fewer references to wrong articles and paragraphs in those justifications.

A key challenge in the Markdown conversion was transforming tables, which were manually reformatted as bullet point lists to preserve content structure.
This step was particularly important, as tables often contain structural elements, e.g., headers, that are vital to accurately interpreting their contained information.
Without transforming tables prior to processing and chunking, essential semantic information could be lost, e.g., when table headers are separated from their corresponding data rows, rendering the latter unintelligible.

\subsection{Implementation}

The RAG system was implemented in Azure AI Foundry, which includes several configurable parameters that can be tuned to optimize performance of the RAG system for a specific task. 
First, the used \emph{embedding model} determines how both the background information stored in the vector database and the user question are converted into vector representations.
The \emph{chunk size} defines the number of tokens in each text chunk stored in the vector database, while the \emph{chunk overlap} parameter controls the extent to which consecutive text chunks overlap, which can affect preservation of context between segments~\cite{azure_ai_search_chunk_2024}.
The \emph{top\_k} parameter specifies how many of the most relevant chunks are retrieved in response to a user query.
Finally, the \emph{efSearch} parameter determines the size of the dynamic candidate list considered during approximate nearest-neighbor search using the HNSW algorithm~\cite{azure_ai_search_hnswparameters_2025}.
We experimented with various RAG configurations and found that, for the task of automated identification of the place of supply in the textbook VAT cases, the best accuracy (93.24~\%) was achieved using the \emph{text-embedding-ada-002} model, a \emph{chunk size} of 1024 tokens, a \emph{chunk overlap} of 0, \emph{top\_k} set to 5, and \emph{efSearch} set to 800.
In this context, ``accuracy'' refers to the percentage of cases where the predicted country (place of supply) matches the expected country in the ground truth.

We first performed a series of experiments to choose the optimal embedding model, and to determine suitable values for the \emph{chunk size} and \emph{top\_k} parameters.
Table~\ref{tab:rag_base_parameters} shows the accuracy of the different RAG configurations, for both the basic preprocessing (\emph{prep-basic}) of the data sources  as well as the extended preprocessing (\emph{prep-ext}) with the removal of cross-references and peripheral elements from the documents.
Interestingly, the removal of cross-references and peripheral elements in preprocessing (\emph{prep-ext}) only had a benefit together with the \emph{text-embedding-ada-002} model, while the \emph{text-embedding-3-large} model showed marginally better accuracy with cross-references and peripheral elements included. 
Since better overall results were achieved using the \emph{text-embedding-ada-002} model, we propose the removal of cross-references and peripheral elements in data preprocessing.
Therefore, we conducted the subsequent experiments to determine the best values for the remaining parameters using the \emph{text-embedding-ada-002} model and extended preprocessing.

Generally, the results with the extended preprocessing (\emph{prep-ext}) reveal a divergence in the performance of the two considered embedding models following preprocessing. 
For the \emph{text-embedding-3-large} model, accuracy generally decreased when using extended preprocessing compared to basic preprocessing (\emph{prep-basic}), indicating that this model's strengths align with tasks requiring a high degree of semantic understanding. 
The simplifications introduced through preprocessing---such
as removing cross-references and citations---may have reduced the complexity of the indexed data, diminishing the relative advantage of this model. 
In contrast, the \emph{text-embedding-ada-002} model showed further improvements, suggesting that its performance is optimized by simpler and better-structured data.

\begin{table}[h!]
\centering
\footnotesize
\caption{Accuracy of RAG configurations for the embedding models \emph{text-embedding-ada-002} and \emph{text-embedding-3-large} with both basic preprocessing and extended preprocessing of the data sources, using different values for \emph{chunk size} and \emph{top\_k}, with \emph{chunk overlap}~= 0 and \emph{efSearch}~= 500. Light gray background color marks the highest accuracy.}
\label{tab:rag_base_parameters}

\bigskip

\begin{tabularx}{\textwidth}{@{}cc|YY|YY@{}}
\toprule
\textbf{chunk size} & \textbf{top\_k} 
& \multicolumn{2}{c|}{\textbf{text-embedding-ada-002}} 
& \multicolumn{2}{c}{\textbf{text-embedding-3-large}} \\
& & \textit{prep-basic} & \textit{prep-ext} & \textit{prep-basic} & \textit{prep-ext} \\
\midrule
64   & 5   & 66.22~\% & 64.86~\% & 68.92~\% & 68.92~\% \\ 
64   & 10  & 66.22~\% & 67.57~\% & 70.27~\% & 67.57~\% \\ 
512  & 5   & 78.38~\% & 79.73~\% & 85.14~\% & 83.78~\% \\ 
512  & 7   & --       & 79.73~\% & --       & 81.08~\% \\
1024 & 3   & 74.32~\% & 82.43~\% & 83.78~\% & 79.73~\% \\ 
1024 & 5   & 81.08~\% & \cellcolor{lightgray}89.19~\% & 83.78~\% & 82.43~\% \\ 
1024 & 7   & --       & 83.56~\% & --       & 79.73~\% \\ 
1500 & 5   & --       & 85.14~\% & --       & 82.43~\% \\  
\bottomrule
\end{tabularx}

\end{table}

With a \emph{chunk overlap} of 0, the documents are split into chunks without overlapping content. 
Increasing this value can help capture contextual information shared between adjacent chunks, potentially enhancing model performance. 
A \textit{chunk overlap} of 10--15~\% of the defined \textit{chunk size} is sometimes recommended~\cite{azure_ai_search_chunk_2024}.
In our experiments, however, we did not find benefits in having the chunks overlap.
As shown in Table~\ref{tab:rag_chunk_overlap}, with the \emph{efSearch} parameter fixed to 500, we found that setting the \emph{chunk overlap} to 50 tokens and 150 tokens, respectively, resulted in reduced accuracy (83.78~\% and 79.73~\%) compared to the baseline accuracy of having no chunk overlap (89.19~\%). 
We attribute the reduced accuracy in case of overlapping chunks to the thus introduced redundancy, which fails to provide meaningful additional context for solving VAT cases while introducing noise into the embeddings and potentially reducing overall effectiveness of the retrieval.

\begin{table}[h!]
\centering
\footnotesize
\caption{Accuracy of the identification of the place of supply using different values for the \emph{chunk overlap} parameter in conjunction with \textit{text-embedding-ada-002}, \textit{chunk size}~= 1024, \textit{top\_k}~= 5, and \emph{efSearch}~= 500. Light gray background color marks the highest accuracy.}

\bigskip

\begin{tabular}{c c}
\toprule
\textbf{Chunk Overlap} & \textbf{Accuracy} \\ 
\midrule
0   & \cellcolor{lightgray}89.19 \% \\ 
50  & 83.78 \% \\ 
150 & 79.73 \% \\ 
\bottomrule
\end{tabular}
\label{tab:rag_chunk_overlap}
\end{table}

We experimented with different values for the \emph{efSearch} parameter, which determines the number of candidate neighbors to be evaluated during retrieval.
Table~\ref{tab:rag_efSearch} shows the accuracy of the identification of the place of supply in textbook VAT cases using different values for \emph{efSearch} in experiments with a fixed embedding model, \emph{chunk size}, \emph{top\_k}, and \emph{chunk overlap}.
Lower values of \emph{efSearch} result in a faster retrieval of relevant chunks but also typically reduce accuracy, as fewer candidate neighbors are evaluated, leading to less precise retrieval.
In contrast, higher values increase the number of considered neighbors, typically enhancing accuracy but also extending computation time.
In Azure AI Foundry, the parameter \emph{efSearch} can be set to a value between 100 and 1000, the default value being 500.
As expected, setting the value of \emph{efSearch} at 200 led to a lower accuracy of 87.84~\% compared to an accuracy of 89.18~\% with the default value of 500.
An increased value of \textit{efSearch} at 800, in turn, improved accuracy to 93.24~\%, demonstrating the benefit of a more comprehensive neighborhood search.
However, further increasing the value of \textit{efSearch} to 1000 led to a decrease in accuracy (83.78~\%).

\begin{table}[h!]
\centering
\footnotesize
\caption{Accuracy of the identification of the place of supply using different values for the \emph{efSearch} parameter, with the \textit{text-embedding-ada-002} model, \textit{chunk size}~= 1024, \textit{top\_k}~= 5, and \emph{chunk overlap}~= 0. Light gray background color marks the highest accuracy.}

\bigskip

\begin{tabular}{c c}
\toprule
\textbf{efSearch} & \textbf{Accuracy} \\ 
\midrule
200  & 87.84 \% \\ 
500  & 89.19 \% \\ 
800  & \cellcolor{lightgray}93.24 \% \\ 
1000 & 83.78 \% \\ 
\bottomrule
\end{tabular}
\label{tab:rag_efSearch}
\end{table}

In addition to fixed-size chunking, with documents segmented into chunks of a predefined token size, we also experimented with \emph{document layout chunking}, where chunk boundaries align with section boundaries in the Markdown documents.
Document layout chunking aims to preserve contextual relationships, ensuring that each chunk corresponds to a meaningful, self-contained unit of information.
However, this method did not improve accuracy in this use case.

\section{Fine-Tuned Large Language Model}\label{sec:finetuning}

In the following, we first describe the fine-tuning dataset before summarizing the fine-tuning process, in particular the optimization of the hyperparameter settings for fine-tuning. 
The implementation of the fine-tuning pipeline and the results of the experiments are available online~\cite{benkel_2026}.

\subsection{Fine-Tuning Dataset}

As a guideline, the use of at least several hundreds of examples representing diverse and representative cases is considered best practice for fine-tuning a pre-trained LLM~\cite{azure_openai_services_customize_2024}. 
In the case of Austrian VAT law, the availability of suitable examples is limited, and many examples are not publicly available. 
Hence, the fine-tuning dataset was compiled from the following sources.

\begin{itemize}
    \item Lecture notes from university courses at the University of Applied Sciences Upper Austria and Johannes Kepler University Linz
    \item Questions from official exams for tax auditors
    \item Skriptum Umsatzsteuer (152) -- Band 1 (in English: \textit{Lecture Notes on Value-Added Tax (152) -- Volume 1}) by Kollmann~\cite{kollmann_umsatzsteuer_2024}    
    \item Skriptum Umsatzsteuer (152) – Band 2: Beispielband (in English: \textit{Lecture Notes on Value-Added Tax (152) -- Volume 2: Casebook}) by Bürgler~\cite{buergler_umsatzsteuer_2024}
\end{itemize}

The selected sources cover a broad range of case studies in VAT law. 
Complex questions containing multiple subquestions were systematically decomposed into individual cases.
In total, the fine-tuning dataset comprised 758 example questions and the corresponding answers.

Fine-tuning of GPT models~\cite{azure_openai_services_customize_2024} in Azure AI Foundry requires the preparation of training data in a \emph{conversation format} corresponding to the specification of the OpenAI Chat Completions API~\cite{openai_completions_2025}. 
The conversation format structures the fine-tuning dataset in the form of messages. 
Each example message in the fine-tuning dataset has content for a \emph{user} role and an \emph{assistant} role.
The content associated with the \emph{user} role corresponds to a question submitted by the user to the AI-based assistant, while the content associated with the \emph{assistant} role corresponds to the expected answer.
Listing~\ref{lst:fine-tuning_data} illustrates the required conversation format of the training data for fine-tuning.
The example snippet is derived from a case in the textbook by Berger and Wakounig~\cite{berger_umsatzsteuer_2022}, which was not part of the actual fine-tuning dataset.

\begin{lstlisting}[
  float=h!,
  basicstyle=\ttfamily\footnotesize,
  caption={Conversation format of the training dataset for fine-tuning, exemplified using a textbook VAT case, translated from German, by Berger and Wakounig~\cite{berger_umsatzsteuer_2022}, which is not part of the actual training dataset.}, 
  label=lst:fine-tuning_data,
  frame=single,
  breaklines=true, 
  breakatwhitespace=true
]
{
  "messages": [
    {
      "role": "system", 
      "content": ""
    }, 
    {
      "role": "user", 
      "content": "Where is the sale taxable? A (Vienna) sells his yacht, which is anchored on Lake Garda in Italy, to B (Graz) in Vienna and hands over all the papers and keys."
    }, 
    {
      "role": "assistant", 
      "content": "According to Section 3 (7) of the Austrian VAT Act, the supply is carried out where the goods are located at the time the power of disposal is obtained. Since the object of purchase (yacht) is located in Italy at the time the power of disposal is obtained, the place of supply is in Italy."
    }
  ]
}
\end{lstlisting}

The fine-tuned model will be used for two different tasks.
The first task is rather narrow and requires the AI-based assistant to determine the place of supply or service provision in textbook VAT cases, which serves to estimate accuracy in the course of hyperparameter optimization and prompt tuning.
The second task is solving a wider range of different types of decision problem for real-world VAT cases beyond the decision of the place of supply or service provision. 
Due to the different requirements of these tasks, two separate system messages are required. 
The respective \emph{system} roles were therefore defined in the corresponding prompt messages and left blank in the fine-tuning dataset.

\subsection{Fine-Tuning Process}

We tuned the hyperparameters for fine-tuning in Azure AI Foundry through an iterative experimentation process that combined automated accuracy evaluation on a dataset of textbook VAT cases with analysis of training/validation curves of both loss and mean token accuracy over the duration of the training (which are available in the online appendix~\cite{benkel_2026}).  
We first compared GPT‑4 and GPT‑4o under default settings to select the better base, which turned out to be GPT‑4o.
In this first development iteration, 110 examples were selected from the fine-tuning dataset.
Both GPT-4 and GPT-4o were fine-tuned on these 110 examples using Azure AI Foundry's default parameter settings, corresponding to a \textit{batch size} of 1, a \textit{learning rate multiplier} of 1, and \textit{epochs} set to 3. 
The fine-tuned models generated during this phase were used in prompt engineering to improve the prompt template (see Section~\ref{sec:prompt_engineering}). 
In addition, these fine-tuned models were used to assess which of the two GPT variants was better suited to the specific task. 
The results on the textbook VAT cases showed that GPT-4o outperformed GPT-4 by a substantial margin.
We thus selected GPT-4o and the prompt message shown in Listing \ref{lst:prompt_basic} to subsequently optimize the fine-tuning hyperparameters, using the entire fine-tuning dataset with 758 examples.

\begin{table}[h!]
\footnotesize
\centering
\caption{Fine-tuning hyperparameter settings and respective accuracy in determining the place of supply in textbook VAT cases. Light gray background color marks the highest accuracy.}
\label{tab:accuracy_ft_configurations}
\bigskip

\begin{tabularx}{\textwidth}{>{\centering\arraybackslash}X *{3}{>{\centering\arraybackslash}X} >{\centering\arraybackslash}X}
    \toprule
    \textbf{Batch Size} & \textbf{Learning Rate Multiplier} & \textbf{Epochs} & \textbf{Accuracy} \\
    \midrule
    1   & 1.0   & 3 & 74.32\% \\
    1   & 1.0   & 4 & 74.32\% \\
    1   & 0.2   & 4 & 78.38\% \\
    1   & 2.0   & 4 & 69.86\% \\
    4   & 1.0   & 4 & 71.62\% \\
    8   & 2.0   & 4 & 81.08\% \\
    16  & 2.0   & 4 & 81.08\% \\
    16  & 2.5   & 3 & \cellcolor{lightgray}85.14\% \\
    20  & 2.8   & 3 & 79.73\% \\
    \bottomrule
\end{tabularx}
\end{table}

To identify the most effective hyperparameter settings for fine-tuning, we split the fine-tuning dataset, using 80~\% for training (606 examples) and 20~\% for validation (152 examples).
During the fine-tuning process, the training metrics (\emph{train loss} and \emph{train mean token accuracy}) indicated how well the model was learning from the training data by measuring prediction errors and token-level accuracy after each step.
The validation metrics (\emph{validation loss} and \emph{validation mean token accuracy}) assessed the model's ability to generalize to unseen data, helping to detect overfitting or underfitting during fine-tuning.
In addition to analyzing loss and token accuracy curves, we also looked at the accuracy of the fine-tuned models in determining the place of supply in the textbook VAT cases.
Table~\ref{tab:accuracy_ft_configurations} shows the accuracy in determining the place of supply for the various fine-tuned models, trained on the 606 training examples using the respective hyperparameter settings.
Early configurations had issues such as instability and overfitting, which were alleviated through gradual tuning, namely, by increasing batch size and learning rate multiplier while limiting the training to three epochs.

Using the default settings initially led to high fluctuation in the loss curves and moderate levels of mean token accuracy over time.
To determine whether increasing training time could improve performance, the number of epochs was set to 4, while all other parameters remained
unchanged compared to the default configuration.
Loss curves then showed patterns similar to those observed in the default configuration, with no steady downward trend, while mean token accuracy  did not substantially improve either.
Furthermore, no improvement in the overall accuracy was achieved.
Hence, we concluded that improvement of model performance would not be achieved simply by increasing the number of epochs.

To reduce fluctuations during training, we lowered the \emph{learning rate multiplier} to 0.2 to achieve a more stable and slower convergence process.
These settings somewhat reduced fluctuations in the loss curves during fine-tuning.
The loss curves, however, remained at a moderate level without showing a clear downward trend. 
Although there was a slight increase in accuracy, the loss and token accuracy curves did not show clear convergence, indicating that the model had not yet reached the optimum training state.

We then set the \emph{learning rate multiplier} to 2 to evaluate the effects of a more aggressive optimization.
With this setting, the loss curves showed a clear downward trend, with the mean token accuracy constantly increasing over the course of the training.
Towards the end of training, however, a divergence between the training and validation metrics became apparent, which suggests overfitting.
Similarly, the final accuracy on the textbook VAT cases was low.

We subsequently set the \textit{batch size} to 4 and the \textit{learning rate multiplier} was set back to the default value of 1. 
This change aimed to reduce the instability of the optimization by averaging the gradients over several examples.
Although the increased batch size contributed to a more stable training process, it cannot considerably improve the generalization capabilities of the model.

We further increased the \emph{batch size} to 8 and the \emph{learning rate multiplier} to 2, while the \emph{number of epochs} remained unchanged at 4. 
These changes were made to improve both the stability and reduce the time to convergence. 
The increased batch size should continue to help smooth the gradients, while the higher learning rate multiplier should speed up the optimization. 
The training loss dropped steadily and reached a plateau towards the end of the training process. 
The validation loss also showed a clear downward trend with only minimal fluctuations, indicating a more stable optimization.
The token accuracy curves reflected this stability.

To further investigate the effects of \emph{batch size}, we increased this parameter to 16, while the \emph{learning rate multiplier} remained at 2 and the \emph{number of epochs} at 4. 
The loss curves then showed a higher degree of stability compared to the previous configurations.
However, reaching the final steps of the training process, a divergence between training and validation losses became apparent. 
This discrepancy can be interpreted as a potential indication of overfitting the training data as the optimization process progresses, which was, however, not supported by the accuracy metrics, since both the training mean token accuracy and the validation mean token accuracy remained stable and increased steadily without large deviations from each other.
Nevertheless, we concluded that a further increase in batch size would not lead to an increase in performance. 
Overall accuracy on textbook VAT cases did also not improve.

In order to avoid overfitting, a reduction in the number of epochs could be a solution.
Hence, we set the \emph{number of epochs} to 3 while increasing the \textit{learning rate multiplier} to 2.5 to speed up convergence.
This adjustment should allow the model to explore larger weighting updates and potentially identify better minima in fewer iterations. 
Both the \textit{train loss} and \textit{validation loss} curves then exhibited a smoother, more consistent downward trend, with fewer oscillations than in previous iterations.
Most importantly, the divergence between training and validation loss was less pronounced, suggesting that the reduction in epochs had successfully addressed the potential overfitting problem. 
The token accuracy curves also showed a steady upward trend. 
The close fit of these curves shows that the model generalizes well to the validation data, further confirming the stability of the training process.

We made further adjustments to \emph{batch size} and \textit{learning rate multiplier} to investigate whether increasing these parameters even more could improve the performance of the model. 
We wanted to find out whether the model could utilize a larger batch size for improved gradient stability and whether a faster learning rate could lead to better minima within the same number of iterations. 
The \emph{training loss} decreased steadily throughout the training process, showing only minimal fluctuations and indicating stable training. 
However, the \emph{validation loss} deviated slightly from the \emph{training loss} towards the end of the training process, suggesting that the higher learning rate and larger batch size may have led to slight overfitting or instability in the validation phase.
The final accuracy on the textbook VAT cases was also lower.

Using a \emph{batch size} of 16, a \emph{learning rate multiplier} of 2.5, and three training epochs---settings, that yielded the best results during the fine-tuning process with 85.14~\% accuracy, stable convergence, and strong generalization---we fine-tuned GPT-4o on all 758 examples from the dataset.
The resulting model was then used in the final evaluation on both the textbook and the real-world cases.

\section{Final Evaluation}\label{sec:evaluation}

Using the best configurations of the AI-based assistant found in previous development iterations, we conduct a final evaluation of the accuracy of the analyses generated for both the textbook cases and the real-world cases.
Note that in the following the prompt template from Listing~\ref{lst:prompt_basic} is used to identify the place of supply in the textbook cases and the prompt template from Listing~\ref{lst:prompt_general_rag} to obtain a general analysis of more complex real-world VAT cases.

\subsection{Textbook Cases}

We evaluate the different configurations of the AI-based assistant in terms of accuracy in determining the correct place of supply in the textbook VAT cases as well as the corresponding justification of the answer.
We manually determined the correctness of the justifications.
We use a configuration that we refer to as \emph{Mocked RAG} as a benchmark to assess the legal reasoning capabilities of the LLM, under the assumption of perfect retrieval of the relevant chunks in the RAG process.
In the \emph{Mocked RAG} configuration, we already provide for each case the relevant chunks of background information, without the need for retrieving those chunks first via a search.
The \emph{Mocked RAG} configuration allows us to answer the following question: \emph{If all of the legal background information that is required to analyze a case is made available to the LLM, what legal reasoning performance can we expect from the LLM?}
The performance of the \emph{Mocked RAG} configuration represents the upper threshold that can be expected from any RAG system using the previously developed prompt message.

Table~\ref{tab:final-textbook-accuracy} shows the accuracy of the different configurations of the AI-based assistant in identifying the correct place of supply in the textbook VAT cases (again as percentage of cases where the predicted country matches the expected country).
The \emph{Mocked RAG} configuration returns the wrong answer for only three cases; using the fine-tuned LLM (\emph{Mocked RAG + FT}) does not improve the performance of \emph{Mocked RAG}.
The high accuracy of the RAG system, returning the wrong answer in only five cases, shows that the retrieval of relevant background information works well.
The fine-tuned LLM and the combination of RAG system with the fine-tuned LLM (\emph{RAG + FT}) shows slightly worse performance than the RAG system alone.

\begin{table}[h!]
\footnotesize
\centering
\caption{Accuracy of identifying the place of supply in the 74 textbook VAT cases using different configurations of the AI-based assistant}
\label{tab:final-textbook-accuracy}
\bigskip
\begin{tabular}{lc}
    \toprule
    \textbf{Configuration} & \textbf{Accuracy}\\
    \midrule
    GPT-4o (plain) & 68.92~\% \\
    RAG + FT & 86.49~\% \\
    Fine-tuned LLM &  89.19~\% \\
    RAG & 93.24~\% \\
    Mocked RAG + FT & 95.95~\% \\
    Mocked RAG & 95.95~\% \\
    \bottomrule
\end{tabular}
\end{table}

\begin{figure}[ht!]
	\centering
	\includegraphics[width=\textwidth]{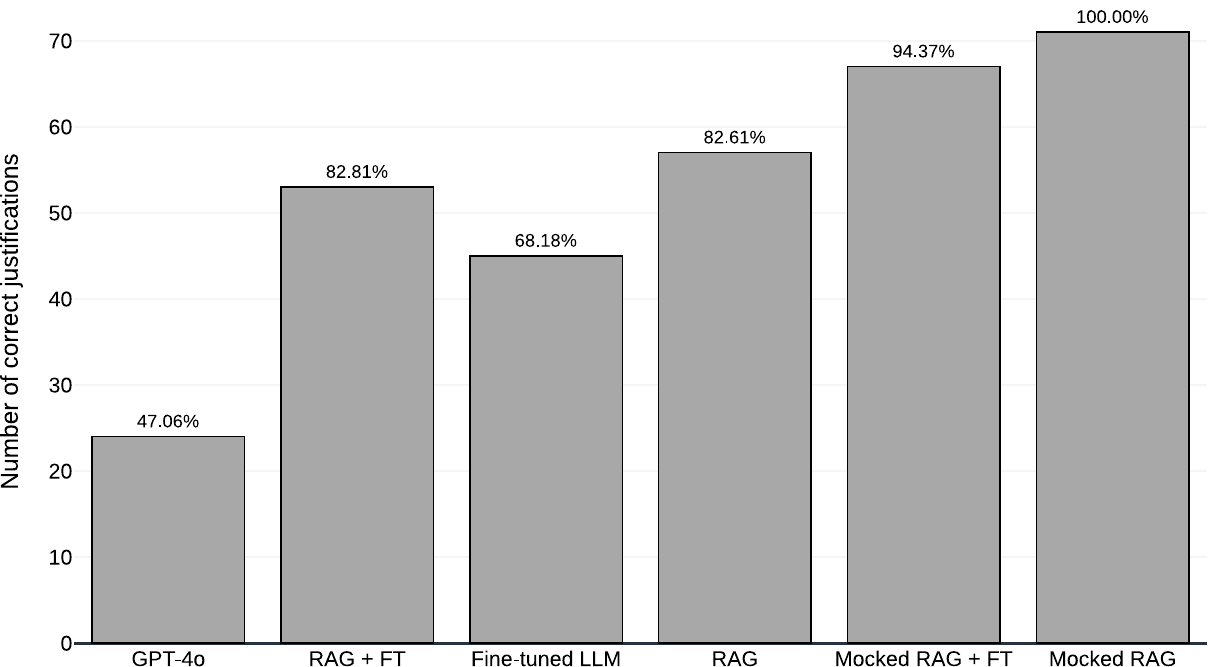}
	\caption{Evaluation of the justifications provided by the best configurations of the AI-based assistant for the answers to the textbook VAT cases where the place of supply was correctly identified}
	\label{fig:final-textbook-justifications}
\end{figure}

Figure~\ref{fig:final-textbook-justifications} shows the accuracy of the different configurations in giving justifications for the identified place of supply; a justification is only considered correct if the identified place of supply is also correct.
In this context, ``accuracy'' refers to the percentage of cases where the justification was manually determined to be correct.
The manual evaluation of the justifications provided for the answers where the place of supply was correctly identified reveals that \emph{Mocked RAG} has all justifications correct, with the combination of \emph{Mocked RAG} and the fine-tuned LLM (\emph{Mocked RAG + FT}) having a slightly worse performance.
The quality of the justifications returned by the RAG system is better than those returned by the fine-tuned LLM.
The use of the fine-tuned LLM does not improve the performance of the RAG system.

An analysis of the mistakes in the justifications revealed two predominant types of mistakes. 
The first type occurred when the justification logically aligned with the applicable legal framework but cited an incorrect legal paragraph. 
In these cases, the reasoning itself was sound, but the specific reference to the legal text was wrong, undermining the overall justification. 
The second type involved justifications that adhered to the reasoning structure of the cited legal paragraph but relied on a paragraph that was irrelevant to the question at hand. 
In such instances, the model demonstrated an ability to mimic the reasoning style of the relevant legal provisions but failed to retrieve the correct paragraph, leading to inaccurate conclusions. 
These errors were particularly common in configurations relying on RAG, the standalone fine-tuned LLM, or their combination.

A more nuanced challenge arose during the use of the fine-tuned LLM.
While this model occasionally produced justifications with matching and correct reasoning as well as legal references, the model often introduced additional contextual information to enhance the comprehensiveness of the responses. 
However, this supplementary information was not always accurate or relevant. 
In such cases, the added details contained factual inaccuracies or hallucinated elements, which rendered the justification incorrect despite its otherwise solid foundation. 
This phenomenon underscores the inherent trade-off between generating more detailed, explanatory responses and maintaining strict adherence to factual and legal accuracy. 
The inability to suppress hallucinations entirely using the fine-tuned model demonstrates a limitation that requires further refinement, particularly in scenarios where precision and reliability are paramount, such as legal reasoning tasks.

A particularly noteworthy observation emerged from the evaluation of combined approaches, such as \emph{RAG + FT} and \emph{Mocked-RAG + FT}.
Tax experts reviewing the justifications noted that the responses generated by these configurations were often exceptionally well-structured, clearly articulated, and highly readable. 
In fact, many of these responses were described as being more elegantly phrased and easier to comprehend than the reference solutions provided in the textbook.
These findings underscore the potential of such combined approaches to generate outputs that are not only accurate but also highly user-friendly---an aspect of particular importance in domains requiring the clear communication of complex legal concepts.
However, that improved linguistic clarity and readability do not necessarily correlate with enhanced factual accuracy. 
Notably, in the case of \emph{RAG + FT}, approximately one-third of the justifications contained incorrect reasoning, resulting in a higher error rate compared to the individual approaches.
This suggests that while combined methods may enhance the presentation and accessibility of legal arguments, they do not inherently mitigate the risk of legal misinterpretations or inaccuracies.

A key insight of the evaluation was the inconsistent nature of the performance of combined approaches compared to standalone configurations.
While the combined \emph{RAG + FT} approach did not consistently outperform either RAG or the fine-tuned LLM individually, there were notable instances where mistakes present in both standalone approaches were successfully resolved in the combined configuration. 
This suggests the presence of a complementary effect, wherein the fine-tuned model's internal knowledge and the external evidence retrieved by RAG occasionally worked in synergy to produce correct answers that neither standalone approach could achieve independently. 
However, this synergistic effect was not systematic, and no clear pattern emerged regarding the resolution or introduction of errors in the combined configurations. 
There were cases in which both standalone approaches produced incorrect justifications, but the combined approach yielded a correct result. 
Conversely, there were also instances in which one standalone approach provided a correct justification while the other was incorrect, with the combined approach either successfully resolving the issue or introducing a new error. 
Furthermore, there were scenarios in which both standalone configurations produced correct justifications, but the combined approach failed to do so, introducing inaccuracies that were absent in the individual methods.

The lack of a recognizable pattern of the performance of the combined \emph{RAG~+ FT} approach indicates that the interactions between the model's internal knowledge, derived from fine-tuning, and the external evidence retrieved by RAG are highly complex and context-dependent. 
The interactions do not follow any predictable logic, suggesting that the combined \emph{RAG + FT} approach introduces an additional layer of complexity. 
This complexity may arise from the model's attempt to reconcile conflicting information or reasoning pathways, particularly when the retrieved evidence does not perfectly align with the fine-tuned model's internal understanding of the domain.

In addition to the configurations with RAG and/or the fine-tuned LLM, we also used GPT-4.1, which has a wider context window than GPT-4o, on the textbook cases while including all legal sources related to the place of supply of goods or provision of services (see Sect.~\ref{sec:data_sources}) in the prompt.
The results show that the model was able to determine the correct place of supply for 60 out of 74 textbook cases, meaning an accuracy of 81~\%, which is below the performance of the other configurations except plain GPT-4o.
An analysis of the justifications shows that 58 of the 74 justifications, i.e., 78.38~\%, were correct, which is below the performance of the \emph{RAG} and \emph{Mocked RAG} configurations but above the performance of the fine-tuned LLM.

\subsection{Real-World Cases}

\begin{figure}[t!]
	\centering
	\includegraphics[width=0.55\textwidth]{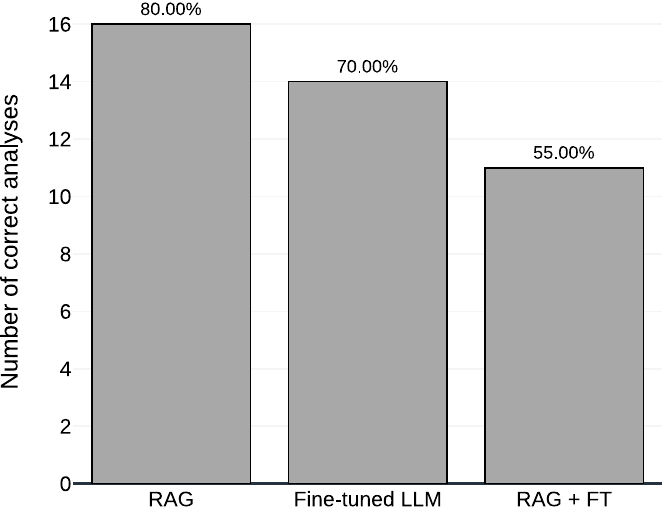}
	\caption{Number of correct analyses for the 20 real-world VAT cases returned by the best system configurations}
	\label{fig:justifications-real-world}
\end{figure}

The evaluation of the analyses generated by the different configurations of the AI-based assistant for the 20 real-world VAT cases reveals notable performance differences. 
The results (Figure~\ref{fig:justifications-real-world}) show that, for the investigated real-world VAT cases, the RAG approach performs best, with 16 correct justifications (80~\%) and four incorrect justifications (20~\%).
The fine-tuned model follows closely with 14 correct justifications (70~\%) and six incorrect justifications (30~\%). 
By contrast, the combination of RAG and the fine-tuned LLM (RAG + FT) has only eleven correct justifications (55~\%), but nine incorrect justifications (45~\%).
In this context, ``accuracy'' refers to the percentage of cases where the returned analysis was factually correct.

We employ McNemar's test~\cite{mcnemar_note_1947} for the analysis of statistical significance of the difference in accuracy between RAG and fine-tuned LLM since McNemar's test is specifically designed to evaluate paired nominal data, where two models (RAG vs. fine-tuned LLM) are applied to the same set of instances (20 real-world VAT cases) and their outcomes are dichotomously classified (correct vs. incorrect).
To mitigate the tendency of McNemar's test to overstate significance in small or sparse paired samples, we apply Yates's continuity correction~\cite{yates_contingency_1934}.
Table \ref{tab:contingency_real-world} presents a $2 \times 2$ contingency table contrasting the accuracy of the analyses of the 20 real-world VAT cases generated by the RAG system and the fine-tuned LLM. 
Of the six cases incorrectly analyzed by the fine-tuned LLM, the RAG system had two correct.
In contrast, of the four cases incorrectly analyzed by the RAG system, the fine-tuned LLM had two correct.
Using McNemar's test with Yates' continuity correction on the experiment data, we obtain a \emph{p}-value of 0.683, which is \emph{not} below the threshold of 0.05 typically required to reject the null hypothesis, meaning that the performance differences between the RAG system and the fine-tuned LLM are not statistically significant.

\begin{table}[t!]
\footnotesize
\centering
\caption{Contingency table for performance comparison of RAG and fine-tuned LLM on the 20 real-world VAT cases}
\label{tab:contingency_real-world}

\bigskip

\begin{tabular}{lcc|c}
\toprule
& \makecell{\textbf{Fine-tuned LLM} \\ \textbf{(Correct)}} 
& \makecell{\textbf{Fine-tuned LLM} \\ \textbf{(Incorrect)}} 
& \textbf{Total} \\
\midrule
\textbf{RAG (Correct)}   & 12 & 4 & 16 \\
\textbf{RAG (Incorrect)} & 2  & 2 & 4  \\
\midrule
\textbf{Total}           & 14 & 6 & 20 \\
\bottomrule
\end{tabular}

\end{table}

An analysis of disagreement cases reveals distinct patterns across the three approaches. 
For \emph{RAG + FT}, all nine incorrect predictions relate to supplies of goods, with six of these involving chain transactions. 
In contrast, \emph{RAG} alone produces four incorrect analyses, only one of which concerns a case that was also incorrectly analyzed by both \emph{FT} and \emph{RAG + FT}. 
Another incorrect analysis concerns a case which was also incorrectly analyzed by \emph{FT} but not by \emph{RAG + FT}, while the remaining two cases were incorrectly analyzed only by \emph{RAG}. 
\emph{FT} alone results in six incorrect cases, five of which coincide with those incorrectly analyzed by \emph{RAG + FT}. 
Notably, all \emph{FT} errors are associated with chain transactions.

Unlike the rather limited task of identifying the place of supply in textbook cases, the rather open-ended task of providing a comprehensive analysis of real-world VAT cases requires a wider range of legal documents.
Since multiple analyses of the real-world VAT cases may be factually correct, with each analysis potentially focusing on different aspects to varying extents, providing an exhaustive list of relevant legal documents is not possible.
Thus, a \emph{Mocked RAG} setup was not considered for the analysis of real-world VAT cases.

We also used GPT-4.1 on the real-world VAT cases while including all legal sources related to VAT (see Sect.~\ref{sec:data_sources}) in the prompt.
The results show that GPT-4.1 with the legal sources in the prompt was able to provide a correct analysis for 13 out of the 20 real-world VAT cases (65~\%), which is below the performance of both the RAG and the fine-tuned LLM configuration, but above the performance of the combination of RAG and fine-tuned LLM.

\section{Discussion}\label{sec:discussion}

In the following, we discuss implications of the findings from our experiments for research and practice.

\subsection{Implications for Research}

The performance of the RAG approach in both textbook and real-world VAT cases indicates its effectiveness in enhancing the LLM's ability to provide accurate justifications in real-world VAT cases. 
By leveraging an expanded document corpus, RAG grounds responses in relevant legal texts, reducing the risk of hallucinations or inaccurate extrapolations.
In contrast, the fine-tuned model, though also effective, does not match RAG's accuracy.
This limitation likely stems from the fine-tuned LLM's reliance on pre-trained knowledge rather than dynamic document retrieval.
While fine-tuning helps the model internalize VAT-specific reasoning patterns, it may struggle to surface case-specific legal references.
In addition, the slightly lower performance may reflect inconsistencies in the fine-tuning data, which, though carefully curated, lacked standardization to a certain degree, potentially hindering the model's ability to generalize across diverse VAT scenarios.

Interestingly, the combination of RAG and the fine-tuned model (\emph{RAG + FT}) does not yield additive benefits but instead underperforms both individual approaches, achieving an accuracy of only 55~\% on the real-world VAT cases.
While such a result may appear counterintuitive at first glance, it suggests that simply combining retrieval-augmented generation with fine-tuning does not guarantee improved performance and may, in fact, introduce adverse interactions between the two approaches.
This underperformance can possibly be attributed to several factors.
First, the integration of retrieved legal texts with pre-learned reasoning patterns can introduce conflicts, especially when the LLM's internal logic diverges from the retrieved content.
These inconsistencies may result in flawed justifications.
Moreover, the added complexity of reconciling retrieved information with learned knowledge may increase susceptibility to errors, particularly in the context of ambiguous or nuanced VAT rules. 
Another possible explanation is that incorporating RAG may reduce the LLM's generalization capability, making the LLM overly reliant on the retrieved documents rather than applying broader legal principles, thus hampering performance when retrieval is suboptimal.

LLMs are able to provide a certain degree of explainability of the returned decisions.
The justifications returned by the AI-based assistant refer to the legal documents that informed the solution, thus providing a kind of audit trail linking the decisions back to the legal documents.
However, these justifications may or may not be correct, the references to legal documents may or may not be correct.
Automatically checking for contradictions is thus not possible.
In this regard, enriching the AI-based assistant with an ontology and a knowledge graph, and requiring the responses to be explicitly linked to the structured knowledge, may facilitate automatic detection of contradictions, which would allow for safe-guarding against wrong decisions being implemented in tax practice.

In tax legal practice, involvement of a human-in-the-loop to assume ultimate authority remains critical.
The modes of human--machine interaction are of crucial interest to research, and may fall on a continuum between ``assisted decision-making, verified decision-making, and delegated decision-making''~\cite{maedche2019}.
The presented AI-based assistant can be considered an example of ``verified decision-making'', rendering decisions but providing human-readable justifications and analyses to allow for a human tax advisor to verify the decisions.

An analysis of disagreement cases between different configurations suggests that chain transactions constitute a particularly challenging pattern, especially for models incorporating fine-tuning, while RAG alone exhibits a more diverse set of mistakes.
A plausible explanation is that chain transactions require multi-step reasoning over relationships between various parties.
Fine-tuning leads a model to rely mostly on learned patterns. 
In the combined \emph{RAG + FT} setting, the bias towards using learned patterns appears to dominate over the retrieved background information, leading to systematic mistakes that are not corrected by the retrieval.

While context windows of LLMs have been widening, the relatively weak performance of long-context GPT-4.1 with all legal sources included in the prompt compared to the RAG-based configurations with narrow-context GPT-4o, particularly \emph{Mocked RAG}, suggests that RAG still has a place in LLM-based systems.
Thus, research into effective, intelligent techniques for retrieval of sources to increase the quality of LLM outputs remains relevant.

\subsection{Implications for Practice}

Regardless of the configuration, the evaluation of justifications revealed several challenges for ensuring the practical applicability of LLMs in real-world VAT advisory contexts. 
A key issue is that tax advisors often build extensive, client-specific knowledge through long-term professional relationships.
Consequently, clients frequently omit crucial details in their inquiries, assuming their advisor already understands the relevant context. 
LLMs, however, lack this implicit background knowledge, which significantly hampers their ability to generate accurate justifications when essential information is missing.
In practice, such background knowledge could be made available in a structured representation, e.g., in form of a knowledge graph.

VAT-related questions are often embedded within complex email exchanges that include attachments such as contracts, financial statements, or other supporting documents.
Human advisors routinely analyze these materials as part of their assessment, incorporating the information into their justifications.
In contrast, the LLMs in our evaluation were not provided with such external documents, leaving them without access to critical contextual data. 
As a result, the models had to generate responses based solely on the textual input received, without the ability to cross-reference legal or financial documents.
In a practical setting, the AI-based assistant should have comprehensive access to relevant email correspondence and other support material, the organization of which poses challenges regarding effective data management.

In real-world practice, an error rate of 20--30~\% in VAT case decisions is unacceptable.
The results show that while LLMs have the potential facilitate automation of legal reasoning to a certain degree, an AI-based assistant for decision-making in tax consulting built entirely on the foundation of LLMs, and by extension RAG or fine-tuning, is insufficient for practice.
Efforts to automate decision-making in tax consulting solely using LLMs must be tempered in light of the results, which underline observations that logic has a fundamental place in legal argumentation~\cite{prakken2015}.
Thus, in tax consulting practice, LLMs should be combined with other mechanisms, e.g., by using LLMs for natural-language processing and initial analysis of a case, while using law-based rules for the final decision-making~\cite{luketina2025}.

Independent of the tools used in legal practice ultimate responsibility has always remained with the tax adviser. 
The AI-based assistant shall serve as support for tax advisers in a way that they use the tool to solve cases more efficiently. 
AI-generated answers have to be checked by tax advisers in a final step. 
The AI-based assistant serves for cross-checking the tax advisor's knowledge and saving time related to the search of legal norms. 
Thus, the tax advisor shall constitute the human-in-the-loop and verify the AI-answers.

Although current prototypes are not ready for fully automated deployment due to the sensitivity of the legal domain, the developed prototypes could serve as valuable support tools for tax professionals, offering initial analyses and helping standardize VAT-related case analysis with further refinement.
The AI-based assistant shall serve as support for a VAT expert, meaning that the AI-based assistant should provide a first solution of VAT cases based on legal norms and legal justifications.
The provided analysis would serve then tax experts as a starting point. 
At the end of the process, a tax expert would have to approve the AI output (human-in-the-loop). 
Such a combination of AI-based assistant as support tool and a tax expert with fundamental VAT knowledge could reduce workload for humans and error rate of the analysis.

From a cost--benefit perspective, the results show that building a RAG-based system is preferable to using a long-context LLM without RAG.
Furthermore, the relatively strong performance of RAG, in particular \emph{Mocked RAG}, suggests that a targeted inclusion of only relevant chunks in the prompt may also have benefits in terms of response quality.

\subsection{Limitations}

This study employs a naïve RAG architecture in which relevant contextual paragraphs for the task at hand are embedded in the prompt to improve correctness of responses and reduce hallucinations~\cite{gao_retrieval-augmented_2024}.
A wide variety of more advanced RAG approaches have been proposed (see~\cite{gao_retrieval-augmented_2024} for an overview).
Although this study only used a naïve RAG architecture, the results of this study are promising in the sense that a comparatively cheap RAG setup can compete with a fine-tuned LLM, which requires additional training effort.

To address the limitation of missing background knowledge on the organization in the experimental setting for analyzing real-world VAT cases, we preprocessed the problem statements presented to the LLMs. 
This preprocessing involved analyzing the tax advisor's responses to extract key information necessary for resolving each case---information often derived from internal knowledge or external documents unavailable to the model. 
While this step helped mitigate the issue of missing context, it was not sufficient to fully close the gap in our experiments.
In some instances, the model had to make assumptions to formulate a response. 
Although these assumptions were not always accurate, justifications were deemed correct if they were legally sound. 
In other words, responses were accepted as valid when they adhered to established VAT principles, even if the assumptions did not perfectly reflect the real case details.
The reliance on assumption-based reasoning underscores a critical challenge of making LLMs work in the daily practice of tax consulting: 
While LLMs can produce technically sound justifications, a lack of access to implicit client knowledge and supporting documents can compromise the accuracy and relevance of their conclusions.

A limitation of this study is that the experiments were conducted using models from a single provider, i.e., OpenAI's GPT family of LLMs.
While this choice allowed for a controlled comparison of RAG and fine-tuning, the findings may not fully generalize to other LLM architectures.
Different LLMs may require different prompting strategies.
The proposed prompt templates were optimized specifically for GPT-4o.
When using other LLMs, the effectiveness of the proposed prompt templates will first have to be confirmed.
Nevertheless, the results of this study may serve as a starting point for the development of solutions using other LLMs.
This study is a systematic investigation of the fundamental suitability of LLMs for reasoning in tax law, which can be further extended to investigate reasoning ability of different LLMs in tax law.

Related work~\cite{dettmers2023} shows that parameter-efficient fine-tuning methods can achieve strong performance with relatively small, high-quality datasets.
Nevertheless, additional samples for fine-tuning may further improve the performance of the fine-tuned LLM.

The absence of significance in the differences in performance between fine-tuned model and RAG may be due to the relatively small sample size.
The selected textbook cases, however, represent distinct structural variations implied by the legal framework, rather than interchangeable observations. 
Consequently, expanding the dataset without introducing genuinely new types of constellations would primarily duplicate existing patterns and would not substantively enhance the analysis, even if it were to lead to statistically significant differences.
The limited number of real-world cases reflects the restricted availability of clearly documented and unambiguous examples within this narrowly defined area. 
The empirical findings should therefore be interpreted with appropriate caution, while recognizing that the dataset achieves comprehensive coverage of the legally relevant scenario space addressed in this study.
Blindly adding new cases without considering their characteristics would not result in a knowledge gain even if it resulted in statistical significance.

\section{Conclusion}\label{sec:conclusion}

In the following, we briefly summarize the findings of our paper and identify potential for future research.

\subsection{Summary}
This study explored the use of LLMs in Austrian VAT law, specifically comparing the performance of two common methods of enriching pre-trained LLMs with task-specific background knowledge---namely, RAG and fine-tuning---in experiments on the analysis of textbook cases as well as real-world cases from a consulting firm.
Optimal configurations for both RAG systems and the fine-tuning process were identified through systematic experimentation of various hyperparameters.
In the experiments, the RAG system consistently outperformed the fine-tuned LLM, although the differences were not statistically significant.
The results suggest that expensive fine-tuning may not be required for LLMs to achieve useful reasoning performance in VAT law.

\subsection{Future Work}

Future work may investigate more advanced RAG approaches.
In particular, Graph RAG~\cite{peng2025} with a structured knowledge base may improve retrieval performance in the primarily logic-based field of tax law. 
In this regard, a promising research direction is the integration of contextual client-specific knowledge into AI-based assistants, potentially through hybrid architectures combining RAG with more structured knowledge graphs.

The growing body of research on \emph{uncertainty quantification}~\cite{hu2023,wagner2024} may inform the development of an uncertainty/reliability score for responses provided by the AI-based assistant for VAT law.
Consequently, a human-in-the-loop could better assess the reliability of the returned response and decide whether to intervene based on the uncertainty/reliability score.

Future work may investigate \emph{retrieval-augmented fine-tuning}~\cite{zhang_raft_2024} for decision-making in tax law.
In such approaches, the model is explicitly trained to consider external documents when answering a question, e.g., by training the model to ignore documents that are not relevant for the specific question.
By jointly optimizing for both retrieval usage and task completion, retrieval-augmented fine-tuning may help address challenges where the combination of RAG and fine-tuned LLM fail to correctly consider the retrieved information, particularly in complex reasoning scenarios that are common in tax law.

Future work may also investigate the design of prompt templates tailored specifically to combinations of a fine-tuned LLM with RAG. 
In particular, prompts could explicitly instruct the model on how to balance retrieved context with the pre-learned knowledge, e.g., by prioritizing external evidence when available while falling back on previously learned patterns only when necessary. 
Such guidance may help mitigate the observed tendency of the model to sometimes not utilize or misinterpret retrieved information, while perhaps improving performance in complex cases that require integrating multiple sources.

Evaluating multiple heterogeneous models would have substantially increased the manual effort required to review legal justifications, which constituted a major bottleneck in the evaluation process.
Building on the presented results, future work will focus on automating the evaluation of AI-generated justifications to reduce reliance on manual assessments by tax experts. 
Developing domain-specific evaluation systems tailored to VAT law would greatly facilitate development of more effective AI-based assistants for tax consulting.
Future work will incorporate AI-assisted evaluation~\cite{gu2024,verga2024} of legal justifications to enable scalable cross-model comparisons.

\section*{Declaration of the Use of Generative AI and AI-Assisted Technologies in the Writing Process}
During the preparation of this work, the authors used ChatGPT by OpenAI, Writefull by Overleaf, and Academic AI by ACOnet for language editing to improve readability and clarity of this work.
The authors used ChatGPT and Academic AI for semi-automated translation of the German example cases and legal sources cited in the paper.
After using these tools/services, the authors reviewed and edited the content as needed and take full responsibility for the content of the published article.

\section*{Acknowledgments}
This work was supported by the grant ``Basisfinanzierung'' of the University of Applied Sciences Upper Austria and constitutes part of the research project ``Use of Large Language Models in the Field of the Austrian Value Added Tax Act: A Comparison of RAG and Finetuning'', which is led by Marina Luketina. The authors gratefully acknowledge use of the services and facilities of the University of Applied Sciences
Upper Austria.

The authors thank ICON Wirtschaftstreuhand GmbH for providing the real-world VAT cases and for conducting the evaluation of the LLM-generated responses.
The authors also thank DATEV for their expert advice and professional support.

\section*{Disclosure Statement}
The authors have no relevant financial or non-financial competing interests to declare.

\bibliographystyle{elsarticle-num}
\bibliography{references}

\end{document}